\documentclass[conference]{IEEEtran}
\usepackage[utf8]{inputenc}
\usepackage[T1]{fontenc}

\usepackage{url}
\usepackage{cite}
\usepackage{hyperref}
\usepackage{amsmath,amssymb,amsfonts}
\usepackage{algorithmic}
\usepackage{graphicx}
\usepackage{textcomp}
\usepackage{xcolor}
\usepackage{subfig} 
\usepackage{tikz} 
\usepackage{varwidth}
\usepackage{stfloats} 
\newcommand\Umbruch[2][1.7cm]{\begin{varwidth}{#1}\centering#2\end{varwidth}}
\usetikzlibrary{shapes.geometric, arrows}
\usepackage[textsize=tiny]{todonotes}
\def\BibTeX{{\rm B\kern-.05em{\sc i\kern-.025em b}\kern-.08em
    T\kern-.1667em\lower.7ex\hbox{E}\kern-.125emX}}

\renewcommand{\baselinestretch}{0.9}

\newcommand*{\ARXIV}{}%

\begin{document}

\title{Pipe Reconstruction from Point Cloud Data}

\author{\IEEEauthorblockN{ Antje Alex, Jannis Stoppe}
\IEEEauthorblockA{\textit{Institute for the Protection
of Maritime Infrastructures} \\
\textit{German Aerospace Center (DLR)}\\
Bremerhaven, Germany \\
antje.alex@dlr.de, jannis.stoppe@dlr.de}
}

\maketitle

\begin{abstract}
Accurate digital twins of industrial assets, such as ships and offshore platforms, rely on the precise reconstruction of complex pipe networks. 
However, manual modelling of pipes from laser scan data is a time-consuming and labor-intensive process. 
This paper presents a pipeline for automated pipe reconstruction from incomplete laser scan data. 
The approach estimates a skeleton curve using Laplacian-based contraction, followed by curve elongation. 
The skeleton axis is then recentred using a rolling sphere technique combined with 2D circle fitting, and refined with a 3D smoothing step. 
This enables the determination of pipe properties, including radius, length and orientation, and facilitates the creation of detailed 3D models of complex pipe networks. 
By automating pipe reconstruction, this approach supports the development of digital twins, allowing for rapid and accurate modeling while reducing costs.

\end{abstract}


\section{Introduction}
In recent years, there has been a growing interest in the concept of digital twins for critical infrastructures \cite{Lampropoulos2024}, as it enhances system security and resilience \cite{Brucherseifer2021}.
Digital twins provide means for real-time monitoring, maintenance and optimisation; disaster simulations and intelligent decision-making support allow more effective responses and mitigation strategies in times of crisis.
They can be utilized for interactive visualisation of complex infrastructures, processes and executive training purposes  \cite{Lampropoulos2024,Cheng2020}. 
In the same way digital twins play a major role in the maritime sector: examples of applications include digital twins for port and terminal facilities, subsea, offshore, and marine pipelines, as well as offshore structures and ships themselves \cite{Bhati2025}.

Creating a digital twin requires a geometrical model as a foundation.
While new industrial sites are often designed and built with accompanying Building Information Modelling (BIM) and Computer-Aided Design (CAD) models, these digital representations may not be available for older sites, incomplete or outdated.
In such cases, alternative methods, such as laser scanning or photogrammetry, may be necessary to collect the current status and create an digital model of the existing infrastructure.
These models, however, lack semantic information, representing only the geometry of the infrastructure without further details on the functions of its parts, thus limiting the use cases of these models to geometric queries.
Manually adding this information complicates the model creation process, so automation represents an opportunity to reduce the cost of a model with improved utility.

While recent advances in machine learning have led to the development of algorithms that can effectively process point cloud data and detect objects within it, it is not sufficient to simply detect the presence of objects for the automated creation of a digital twin.
Rather, it is essential to recognize and extract object-specific properties, such as shape, size, orientation and connections to other objects. 

The main objective of this research is to create an accurate pipe reconstruction from incomplete point  cloud data, as pipes are frequently occurring industrial objects.
A specific focus is set on determining precise geometric and functional parameters, as those are crucial  for generating accurate digital twins, which are essential in various industrial and maritime applications. 
To achieve this goal an algorithm based on Laplacian-based skeleton contraction was developed, 
incorporating elongation and smoothing operations in its refinement step to generate accurate parameter estimates.

\section{Related Work}
To add semantic information to the point cloud data, it needs to be properly segmented, meaning each point in the data set that represents a class of interest should be assigned to this class (e.~g.\ all points belonging to pipes should be assigned to the pipe class).
This can be achieved through the application of machine learning techniques \cite{Cheng2020} or manual work \cite{Ziamtsov2021}, \cite{Lee2013}.
Additionally, some approaches, such as \cite{Kawashima2014} and \cite{Qiu2014}, remove non-relevant data by similarity analysis.

In \cite{Qiu2014} point cloud data is segmented based on cylinder orientation using global similarity.
To identify the principal cylinder directions, point normals and perpendicular directions are projected onto a Gaussian sphere and then clustered.
With the principal cylinder directions identified, the associated points and normals are mapped onto a 2D plane.
In this representation the circle centers form again clusters, which are then separated into individual pipe segments.
Finally, by finding the intersection of principle axis junctions and bends are reconstructed.

Normal-based region growing is applied to segmented the point cloud into regions of neighbourhood and similar normals in \cite{Kawashima2014}.
For each region the eigenvalues and corresponding eigenvectors are determined.
The points within a region are then projected onto a plane with the same normal vector as the main eigenvector, and a circle is fitted to the point set.
A threshold radius is used to determine whether the points belong to a pipe segment or to other objects.
As the main eigenvector of a junction or elbow does not correspond with the main pipe axis, those parts were subdivided further in an iterative way. Through this process, the algorithm segments the point cloud into straight pipes, elbows, junctions, and non-pipe objects.
Following segmentation,  the pipe radius $r$ and axis of each region is refined by a second 2D circle fit.
To connect individual pipe segments, a sphere with a radius of 1.2 times the pipe radius is used to trace the axis of each segment.
The sphere is moved along the segment's axis, and if points from a different segment are detected within the sphere, the two regions are merged and the sphere continues to trace the axis of the second region.
This process is repeated multiple times to identify any untraced regions, which are then added to the previous axes.

In \cite{Cheng2020} non-relevant point cloud data is removed using a convolutional neural network (CNN).
The CNN not only segments the point cloud into different part types (pipe, flange, elbow, tee, cross and none), but also predict the pipe radius from a set of 23 discrete values and determines the pipe orientation.
After the segmentation, points belonging to individual pipe parts are clustered using DBSCAN \cite{Ester1996}.
The resulting clusters are then used to construct a connected graph, which represents the relationships between the different pipe parts.
In a final step, the graph nodes are replaced with predefined 3D part models.

If the point cloud consists only of pipes, a skeleton can be directly derived from the point cloud.
In \cite{Lee2013} a Voronoi diagram is used to determine skeleton candidates, which represent points along the central axes of the pipes.
To create a connected skeleton graph from these candidates, a Laplacian-based approach from \cite{Cao2010} is used.
Pipe parts can be identified then by the relationship between neighbouring nodes.
For example, tees are identified at nodes with more than two connected neighbours.
Elbows, on the other hand, are detected when the angle between a node and its two neighbouring nodes exceeds a predetermined threshold angle.
Finally, the pipe radius of each segment is estimated by calculating the median distance between the k-nearest neighbours of the skeleton and surface points.

Not pipes but complex tubular structures are reconstructed in \cite{Liu2022}. Using Laplacian-based contraction \cite{Cao2010} on the point cloud itself, a first skeleton graph is estimated.
In a second step the skeleton is refined and centred using a rolling sphere algorithm.
The central axes of individual structures are segmented using a region growing approach that involves testing the similarity of linear direction vectors between neighbouring points.
The tubular component radius and centre axis points are determined by 2D circle fitting of component slices. Based on that, the geometric parameters of the tubular components are modelled in Autodesk Revit.

Reconstruction of tubular components is also applied in other fields, such as tree and plant reconstruction.
In the context of tree reconstruction, \cite{Meyer2023} builds upon the Laplacian-based skeletonization approach of \cite{Cao2010} by incorporating semantic information. 
The process begins with the segmentation of the tree point cloud into trunk and branch points using a Deep Graph Convolutional Neural Network (DGCNN). 
The semantic information obtained from this step is then integrated into the contraction step of the skeletonization algorithm, resulting in a significantly improved skeleton, particularly at the transition points between the trunk and branches.

A method to improve the skeleton behaviour at branching points of plants is proposed in \cite{Ziamtsov2021}. 
The approach involves constructing a skeleton graph by subdividing the point cloud into levels of neighbouring points, based on their distance from the root.
The centres of these levels are then connected to construct the initial skeleton.
To refine the skeleton at junction regions, the point normals and perpendicular directions are projected onto a Gaussian sphere and clustered, similar to the approach described by \cite{Qiu2014}. 
The intersection point of the clustered direction vectors is used to determine the refined skeleton point, resulting in a more accurate representation of the plant's branching structure.

\section{Methods and Material}

\subsection{Data set generation} 
Pipe reconstruction data can be generated in different ways.
One is to use real scanned point clouds of pipe systems where the point cloud is either labelled manually \cite{Lee2013}, \cite{Kawashima2014}, \cite{Agapaki2022}, \cite{Cheng2020} or by using information from BIM and CAD data \cite{Kawashima2014}, \cite{Liu2022}.
Furthermore virtual scans of 3D models are used to generate synthetic point clouds \cite{Cheng2020}.

There is no large database of labelled industrial facility point clouds \cite{Agapaki2019}, especially not with further semantic information.
Hence, a synthetic data set was created.

\definecolor{blue}{RGB}{0,153,206}
\definecolor{black}{RGB}{0, 0, 0}
\definecolor{white}{RGB}{255, 255, 255}
\definecolor{green}{RGB}{156,192,69}
\definecolor{gray}{RGB}{135,135,135}
\tikzstyle{nodeblue} = [rectangle, minimum width=1cm, minimum height=1cm, text centered, font=\small, color=white, draw=blue, line width=1, fill=blue, rounded corners, outer sep=0.1cm]
\tikzstyle{nodegreen} = [nodeblue, draw=green, fill=green]
\tikzstyle{nodegray} = [nodeblue, draw=gray, fill=gray]
\tikzstyle{arrow1} = [thick, draw=black, line cap=round, line width=2, ->, >=stealth]
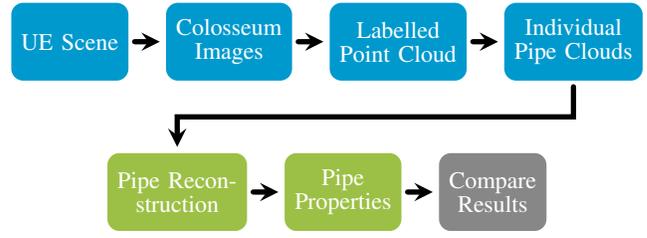
\begin{figure}
	\centering
	\begin{tikzpicture}[node distance=2cm]
		\begin{scope}[]
			\matrix[column sep=0.5cm]{
				\node (node1) [nodeblue] {UE Scene}; &
				\node (node2) [nodeblue] {\Umbruch{Colosseum Images}}; &
				\node (node3) [nodeblue] {\Umbruch{Labelled Point Cloud}}; &
				\node (node4) [nodeblue] {\Umbruch{Individual Pipe Clouds}}; \\
			};
		\end{scope}
		\begin{scope}[yshift=-2cm]
			\matrix[column sep=0.5cm]{
				\node (node5) [nodegreen] {\Umbruch{Pipe Reconstruction}}; &
				\node (node6) [nodegreen] {\Umbruch{Pipe Properties}}; &
				\node (node7) [nodegray] {\Umbruch{Compare Results}}; \\
			};
		\end{scope}
		\draw [arrow1] (node1) --  (node2);
		\draw [arrow1] (node2) --  (node3);
		\draw [arrow1] (node3) --  (node4);
		\coordinate (c1) at ([yshift=-1.0cm]node4);
		\draw [arrow1] (node4.south) -- (c1) -| (node5.north);
		\draw [arrow1] (node5) --  (node6);
		\draw [arrow1] (node6) --  (node7);
	\end{tikzpicture}
	\caption{Process of pipe point cloud generation (blue) and pipe reconstruction (green)}
	\label{fig:process}
\end{figure}

To create the required data without the need for time-consuming, manual labelling, the Unreal Game Engine (UE) \cite{Unreal} and Colosseum \cite{Shah2017}, which is an open source simulator for autonomous robotics built on UE were used.
The general overview of data generation is visualized in Figure~\ref{fig:process}.
In the first step, a pipework environment has to be created in UE, either manually or automatically by using a component catalogue as in \cite{Kai2023}.
Pipes are generally generated by extruding a circle along a spline, and thus are defined by an outer radius, spline points and spline tangents.
Using that definition, pipes follow a continuous curve and thus, in this data set, do not have junctions and are not semantically separated into their distinct parts.

Using Colosseum \cite{Shah2017} and the workflow of \cite{Alvey2021} a virtual camera is positioned at multiple locations in the scene and rotated horizontally, mimicking a laser scan.
The virtual camera captures RGB, depth and segmentation images.
No manual labelling is needed, as the segmentation is done based on the object ID (= name) in the scene.
If the objects are named properly, images and thus point cloud data is segmented based on semantics or instance.
Based on the images, camera position and properties a 3D point cloud can be created based on the pin hole camera model.
Similarly to physical laser scanners, the resolution of the point cloud and the shadowing and occlusion can be improved by increasing the resolution or number of camera positions as well as decreasing the turning interval.

\subsection{Data set} 
The scene to generate our data set is an industrial indoor site, build from the component catalogue described in \cite{Kai2023}.
Besides other components pipes of different radii (6-25~cm) and spline length (0.5-22.5~m) can be found.
For capturing a point cloud of the scene, six camera positions, each with 36 images (10\textdegree~turning interval, resolution $256 \times 192$) were used. 
\textit{Pipe}-class points are filtered from the point cloud and divided based on the pipe instance, so point clouds of individual pipes are created.
Due to the pipe generation process, the individual pipe instances do not have any junctions.
The data set consists of 51 pipe objects, of which 15 are bends and 36 are pipes of different complexity.
Additionally to the point cloud data, the ground truth pipe properties, i.~e.\ outer radius, spline points and tangents, are available from the engine and can be used to evaluate the reconstruction process.

\subsection{Reconstruction Method} 
\begin{figure*}
\centering
\includegraphics[width=\textwidth]{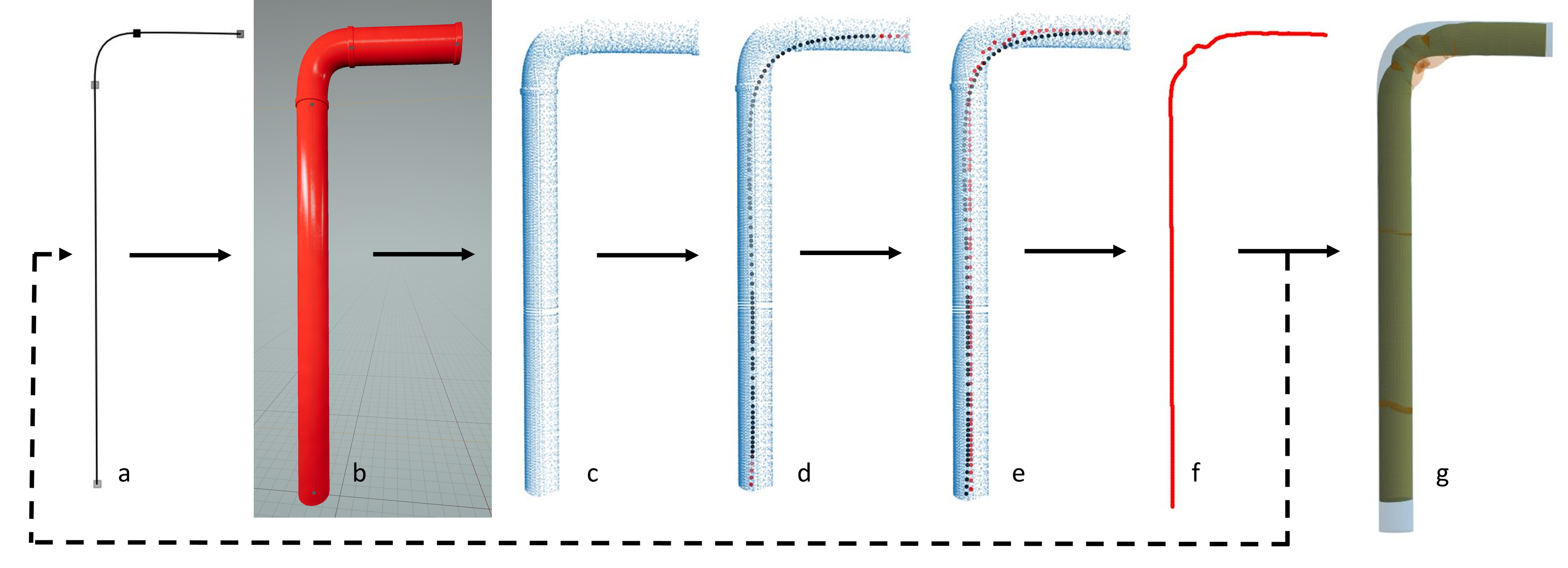}
\caption{Pipe reconstruction process: Spline points and tangents (a) serve as a foundation to build a pipe in UE (b).
Using Colosseum, a point cloud is generated (c).
A skeleton (d) without junctions (black) is generated and elongated (red).
Rolling Sphere Algorithm shifted skeleton points (e, red) are enhanced via 3D smoothing (f).
(g) illustrates the intersection over union (IoU) of the reconstructed pipe volume between ground truth  (blue) and reconstructed pipe hull (orange).}
\label{fig:reconstructprocess}
\end{figure*}
The overall pipe reconstruction process is shown in Figure~\ref{fig:reconstructprocess}~d-g) and contains the following process steps:
\begin{itemize}
	\item Skeletonization (Figure \ref{fig:reconstructprocess} d)
	\item Longest path (Figure \ref{fig:reconstructprocess} d)
	\item Skeleton elongation (Figure \ref{fig:reconstructprocess} d)
	\item Rolling Sphere Algorithm (Figure \ref{fig:reconstructprocess} e)
	\item 3D Curve Smoothing (Figure \ref{fig:reconstructprocess} f)
	\item Hull reconstruction (Figure \ref{fig:reconstructprocess} g)
\end{itemize}

\paragraph{Skeletonization}
Initially, the point cloud $P$ is simplified into a curve skeleton (see Figure \ref{fig:reconstructprocess} d), a compact 1D representation that captures the fundamental topological characteristics of a point cloud while preserving its essential shape.
The point cloud is collapsed to a zero-volume point set using Laplacian-based contraction \cite{Cao2010} by iteratively solving the linear system:
\begin{equation}
\label{Laplacian}
\left[\begin{matrix} W_L L \\ W_H \end{matrix}\right] P' = \left[\begin{matrix} 0 \\ W_H P \end{matrix}\right]
\end{equation}

\begin{figure}[b] 
    \centering
  \subfloat[]{\includegraphics[height=190pt]{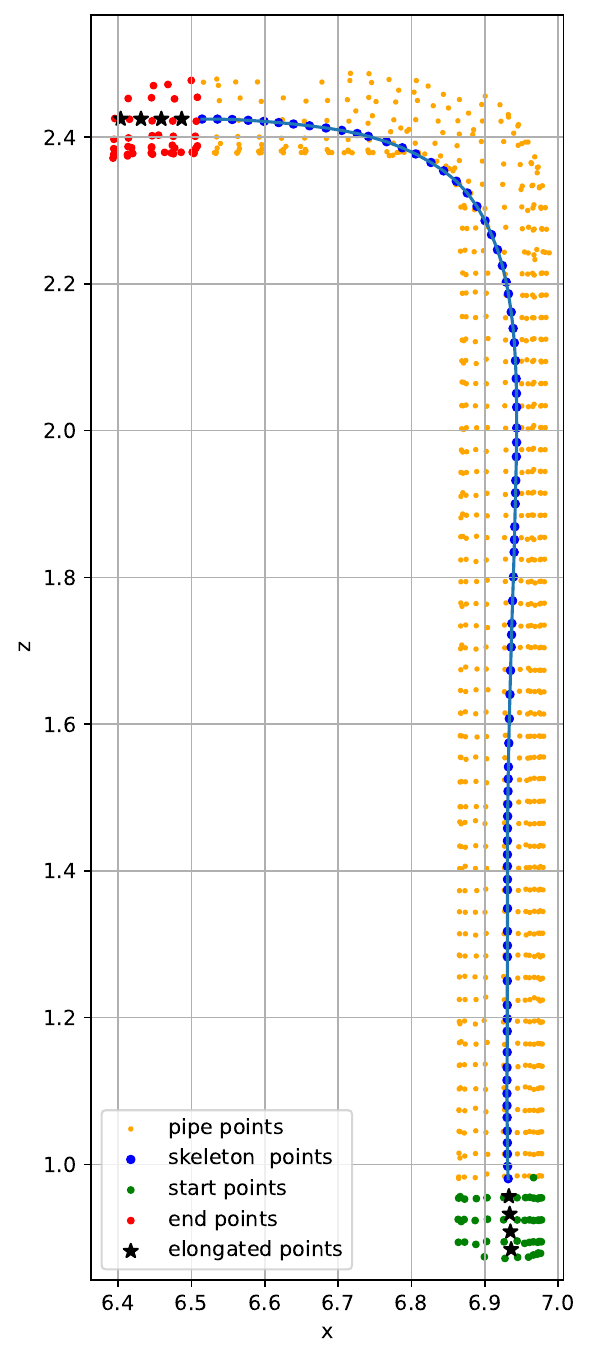}}
    \hfill
  \subfloat[]{\includegraphics[height=190pt]{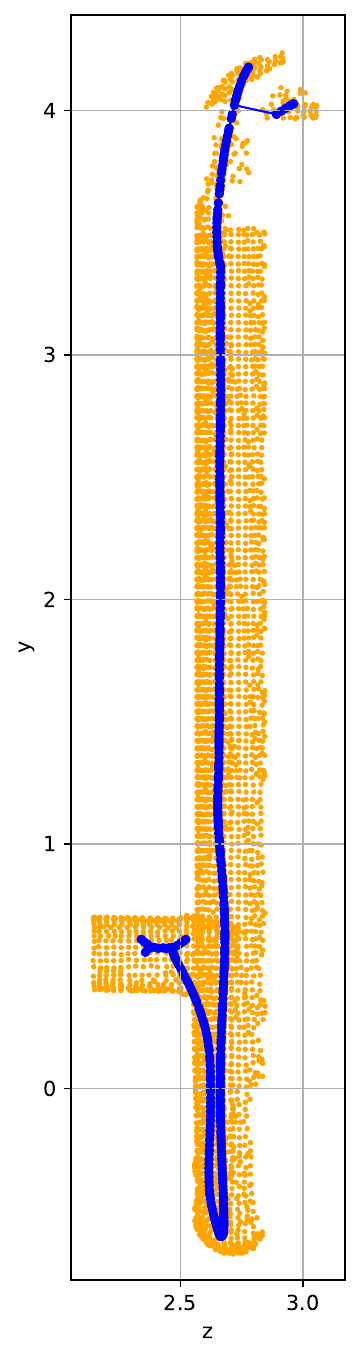}}
    \hfill
  \subfloat[]{\includegraphics[height=190pt]{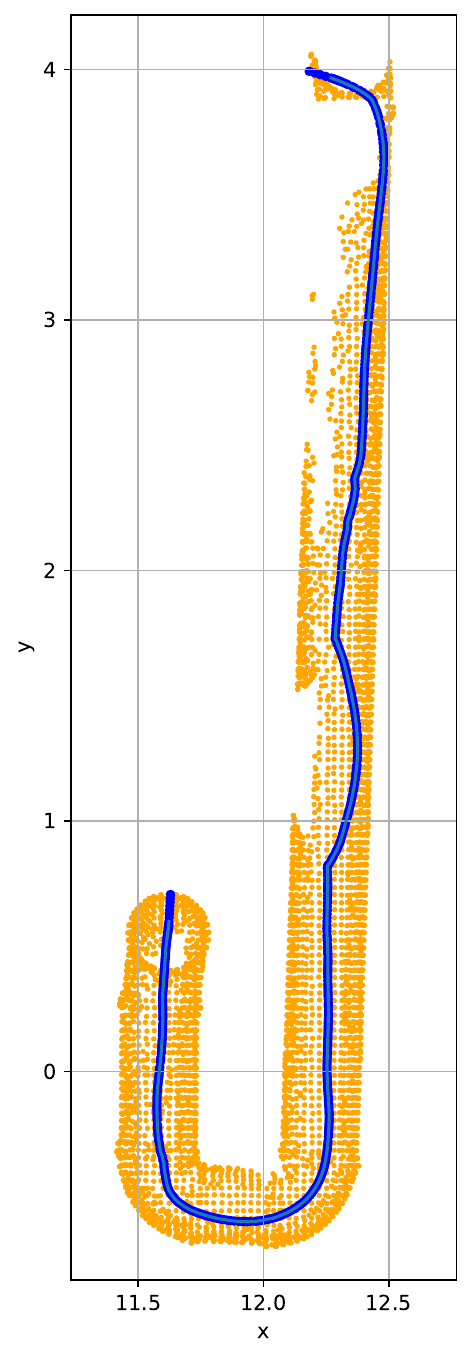}}
  \caption{(a) shrunken skeleton, (b) non continuous skeleton with junctions, (c) skeleton out of middle axis }
  \label{fig:skeleton_effects} 
\end{figure}

$L$ is a cotangent Laplacian matrix, $P'$ is the contracted point cloud, and $W_L$ and $W_H$ are diagonal weight matrices controlling the effect of contraction and attraction.
In each iteration $t$, $W_L$ and $W_H$ are updated, and $L$ is recalculated.
To convert the contracted point set $P'^t$ into a graph, an initial connectivity between the points in $P'$ is build based on farthest-point sampling and a sphere of given radius.
Redundant edges are removed iteratively by contracting the shortest edge to its midpoint and deleting any adjacent triangles.
Using the implementation of \cite{Meyer2023}, this skeletonization approach is performing well on plain point cloud data, even if the point cloud is incomplete \cite{Cao2010}, which is a main issue in the used data set.
Three effects can be observed:
 \begin{itemize}
    \item due to the contraction, the skeleton is shorter than the original pipe (Figure \ref{fig:skeleton_effects} a)
    \item the skeleton does not form a continuous line (Figure \ref{fig:skeleton_effects} b)
    \item the skeleton does not indicate the original middle axis of the pipe, but is closer to the side of the surface where there are more points (Figure \ref{fig:skeleton_effects} c)
 \end{itemize}

\paragraph{Longest path}
The derived skeleton may contain branches and loops (Figure \ref{fig:skeleton_effects} b), which can be problematic for further processing.
To address this, a continuous spline route is created that accurately represents the pipe's geometry.
Given that the data set does not contain any junctions, incorrect branches can be eliminated by analysing the connectivity of the skeleton points.
Specifically, the approach involves identifying the shortest paths between all skeleton points that have only one neighbouring point.
These points are essentially the ``endpoints'' of the branches.
By finding the shortest paths between them, it is possible to determine which branches are the most direct and likely to represent the actual pipe geometry. Once the shortest paths have been identified, the longest path among them is selected as the final skeleton of the pipe.
This path is considered to be the most representative of the pipe's geometry, as it spans the greatest distance and is likely to capture the most significant features of the pipe. 

\paragraph{Elongation}
To counteract the shortening effect caused by the contraction process (Figure \ref{fig:skeleton_effects} a), the longest route obtained from the previous step is subsequently elongated at both its start and end points.
The direction of elongation is determined by the first and last two points of the skeleton, which define the orientation of the pipe at its start and end points, respectively.
To determine the extent of the elongation, a set of points, located within the elongation direction, is identified within a maximum distance (here 0.5~m) from the start and end points of the skeleton.
Those points belonging to the start area are highlighted in green, while those belonging to the end area are highlighted in red.
The largest distance between the start or end point and each point in the corresponding selection defines the amount of extension required for the spline along the predetermined direction.
Finally, equidistant spline points (2~cm intervals) are added to the skeleton spline in the elongation direction, effectively extending the pipe's geometry to its original length.
 
\paragraph{Rolling Sphere Algorithm}
Due to occlusions caused by other objects or limitations in the scanning process, such as when a pipe is located in front of a wall and can only be scanned from one side, the resulting point cloud in incomplete and contains holes. 
These holes have a significant impact on the derived skeleton, as illustrated in Figure \ref{fig:skeleton_effects} c.
To overcome these, the skeleton points are shifted towards the central axis of the pipe applying a rolling sphere algorithm according to \cite{Liu2022}. 
Assuming an ideal scenario where the central axis and the skeleton axis of the full tubular component coincide, a sphere with a radius $r$, equivalent to the radius of the tubular component is translated along the central axis.
As it intersects with the component's surface, it forms a complete ring.
Projecting the points on the ring into a plane with the skeleton axis direction (direction between two neighbouring skeleton points) as normal, a 2D circle can be fitted into it.
The circle's properties, i.~e.\ radius $r$ and center point coordinates, can be determined.
Projecting the circle centre coordinates back in 3D, they coincide with the skeleton point.
However, in cases of incomplete point clouds, the skeleton axis does not align with the central axis, and thus the intersection between the tubular component's surface and the sphere with radius $r$ will only result in a partial ring.
Nevertheless, after 2D projection a circle can be fitted into the point set, from which the radius and circle centre can be determined in the same way as in the ideal case.
Back projection of the circle centre in 3D results in a corrected skeleton point, shifted towards the pipe axis, correcting the error introduced by an incomplete point cloud.
In contrast to \cite{Liu2022} using a sphere of fixed and known radius, here a variable growing sphere is used.
If less than $m$ points are inside a sphere with radius $r$ and a skeleton point $s$ as centre point, the radius is increased iteratively until more than $m$ points are inside the sphere or a maximum number of iterations is reached.
If growth is truncated due to too few points, the skeleton point is deleted and the algorithm continues with the next skeleton point.
If enough points are detected inside the sphere, those points are projected along the direction axis, forming a circle shape in 2D.
Using RANSAC \cite{Ransac}, the centre point $c$ of the (incomplete) circle and its radius $r_*$ is determined.
The corrected skeleton point $s_*$ is the back projection of the circle center $c$.
For the next skeleton point $r_*$ is used as starting sphere radius.
It might occur that multiple circle centre points are back projected on the same corrected skeleton point $s_*$.
To get a continuous and non-redundant skeleton, duplicated skeleton points are removed from the graph.

\paragraph{3D Curve Smoothing}
Smoothing of the skeleton is done according to \cite{Santana2021} by solving the evolution equation 
\begin{equation}
\label{evolution_equ}
\mathcal{C}_t = -\nabla d_{\partial A} \left( \mathcal{C} \right) + wk\mathcal{N},
\end{equation}
where $\mathcal{C}_t$ represents the smoothed 3D curve at iteration step $t$ of the initial skeleton curve $A$, $\mathcal{N}$ is the normal, $k$ the curvature and $w$ is the weight of smoothing.
With a larger value of $w$ the regularization effect becomes stronger. 

In a first step, the ordered skeleton is reparameterised to an equally spaced point set.
From that a 3D image representing the voxelised volume of the skeleton is built.
Similar to the Dijkstra's algorithm \cite{Dijkstra1959}, the distance between neighbouring skeleton points is computed in a second step.
The curve is smoothed by minimising the energy, which is dependent on the weight, the curve length and the distances.
Reparameterisation and energy minimisation is repeated iteratively until the gradient descending method reaches a local minimum.
Due to the fine voxelisation, the number of skeleton points after reconstruction is up to 500 times higher than the required number of spline points, used in UE.

\paragraph{Reduce Spline Points}
To reduce the amount of spline points, the Ramer-Douglas-Peucker (RDP) algorithm \cite{Ramer1972, Douglas1973} is applied to the ordered set of spline points. 
The algorithm relies on a distance parameter $\epsilon$, set to $0.6\overline{r}$. 
The first and last point in the set are connected via a line. 
The distances between the line segment and every point are calculated. 
The farthest point is choosen. 
If the distance is greater then $\epsilon$ the point is kept, otherwise it is discarded from the point set. 
This process is then recursively applied to the two new line segments between the kept point and the start or end point, respectively.

\paragraph{Hull Reconstruction}
While the spline points and the radius will later be used to recreate the pipes in UE, a circle with the mean radius $\overline{r}$ is extruded along the spline, creating the mesh of the pipe. 

\section{Results}
The reconstruction process provide a skeleton graph including spline point positions, axis direction, pipe radius and axis length. 
The ground truth data is represented by spline points, spline tangents and pipe radius.
From the pipe parameters of ground truth data and reconstructed data a mesh of the pipe surface and volume is derived.
The reconstruction method is evaluated on the intersection of union (IoU) of pipe volumes, the pipe radius $r$ and the pipe axis length $l$.
The ground truth data may include invisible parts of pipes inserted into each other. 
Furthermore parts may not be covered in the simulated laser scan due to occlusion. 
Thus the comparison is limited to visible parts only, defined by bounding volumes according to the point cloud boundaries.

\begin{table}
\caption{IoU for different reconstruction techniques}
\label{tab:IoU}
\centering
\begin{tabular}{lrrr}
                 & IoU             & IoU w/o bends   & IoU w/ bends\\
\hline\\[-0.7em]
pure skeleton    & 0.2958          & 0.3907          & 0.0680\vspace{0.5em}\\
base skeleton    & 0.5754          & 0.7476          & 0.1622\\
+ smooth         & 0.5775          & 0.7515          & 0.1598 \\
+ elong          & 0.6475          & 0.7544          &  \textbf{0.3909}\\
+ elong + smooth & \textbf{0.6497} & \textbf{0.7591} & 0.3871\\
\hline
\end{tabular}
\end{table}

The IoU for different reconstruction steps are shown in Table~\ref{tab:IoU}, with pure skeleton referring to the reconstruction based on the longest spline path without junctions and base skeleton to the reconstruction via the centered spline axis after the rolling sphere algorithm. 
Notice that $\overline{r}$ for the pure skeleton is taken from the rolling sphere algorithm, and is thus identical to the mean radius used for the base skeleton.
Using the centred axis, the IoU is nearly doubled from 0.2958 to 0.5754 from pure to base skeleton, independent of the kind of pipe complexity.
IoU increases by 0.07 when adding the elongation to the base skeleton.
The greatest effect can be seen for bends, where it is more than doubled.
Smoothing has only a minor effect on IoU, but produces a subjectively nicer mesh.

The mean radius is always underestimated.
As it can be seen in Table~\ref{tab:r_ratio}, smoothing and elongation has no significant effect on $\overline{r}$. 

\begin{table}
\caption{Results for radius determination}
\label{tab:r_ratio}
\centering
\begin{tabular}{lrrr}
 & $\overline{r}/r_{gt}$ & $\overline{r}/r_{gt}$ w/o bends & $\overline{r}/r_{gt}$ w/ bends\\
 \hline\\[-0.7em]
pure skeleton$^*$ & 0.8658 & 0.9354 & 0.6988\vspace{0.5em}\\
base skeleton     & 0.8658 & 0.9354 & 0.6988  \\
+ smooth          & 0.8658 & 0.9354 & 0.6988 \\
+ elong           & 0.8796 & 0.9280 & 0.7634 \\
+ elong + smooth  & 0.8796 & 0.9280 & 0.7634 \\
\hline\\[-0.7em]
\multicolumn{4}{l}{\begin{footnotesize}$^*$ $\overline{r}$ is taken from  rolling sphere algorithm\end{footnotesize}}
\end{tabular}
\end{table}

The length of the pipe axis is the sum of shortest distance between two neighboring spline points.
Because ground truth data consist of the smallest possible number of spline points, additional spline points are added to cover the length in bending areas more thoroughly.
The absolute length ratio is shown in Table~\ref{tab:l_ratio}.
Length is both under- and overestimated.
The absolute value of $l/l_{gt}$ is not significant increased by smoothing, but it ensures that the length of the pipe is no longer overestimated. 

\begin{table}
\caption{Results for pipe length determination} \label{tab:l_ratio}
\centering
\begin{tabular}{lrrr}
 &  $l/l_{gt}$ & $l/l_{gt}$ w/o bends& $l/l_{gt}$ w/ bends \\
 \hline\\[-0.7em]
pure skeleton & 0.6837 & 0.8800  &  0.3458\vspace{0.5em}\\
base skeleton & 0.7021 & 0.8771 & 0.4156  \\
+ smooth & 0.7099 & 0.8910 & 0.4084  \\
+ elong & 0.8486 & 0.9044 & \textbf{0.8480}  \\
+ elong + smooth & \textbf{0.8587} & \textbf{0.9271} & 0.8280 \\
\hline
\end{tabular}
\end{table}

\begin{figure*}{} 
    \centering
  \subfloat[]{\includegraphics[width=0.30\linewidth]{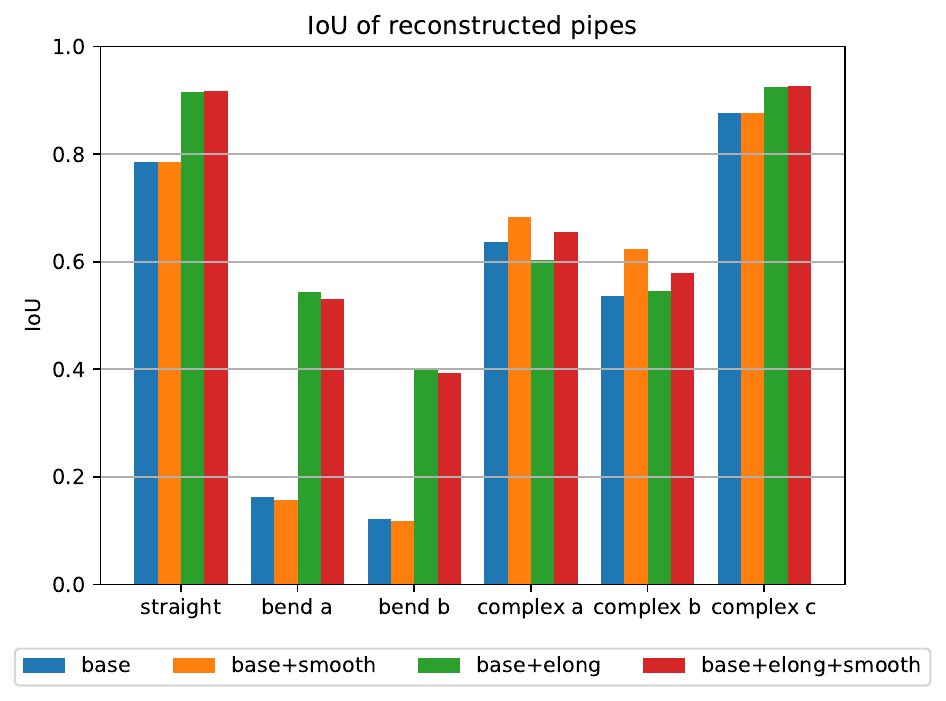}}
    \hfill
  \subfloat[]{\includegraphics[width=0.30\linewidth]{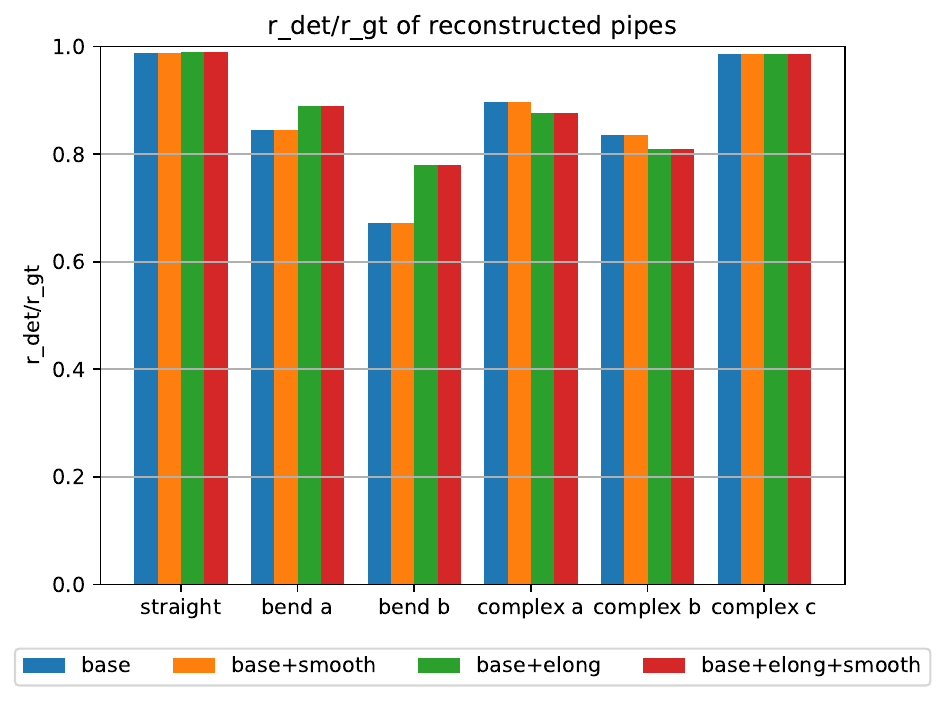}}
    \hfill
  \subfloat[]{\includegraphics[width=0.30\linewidth]{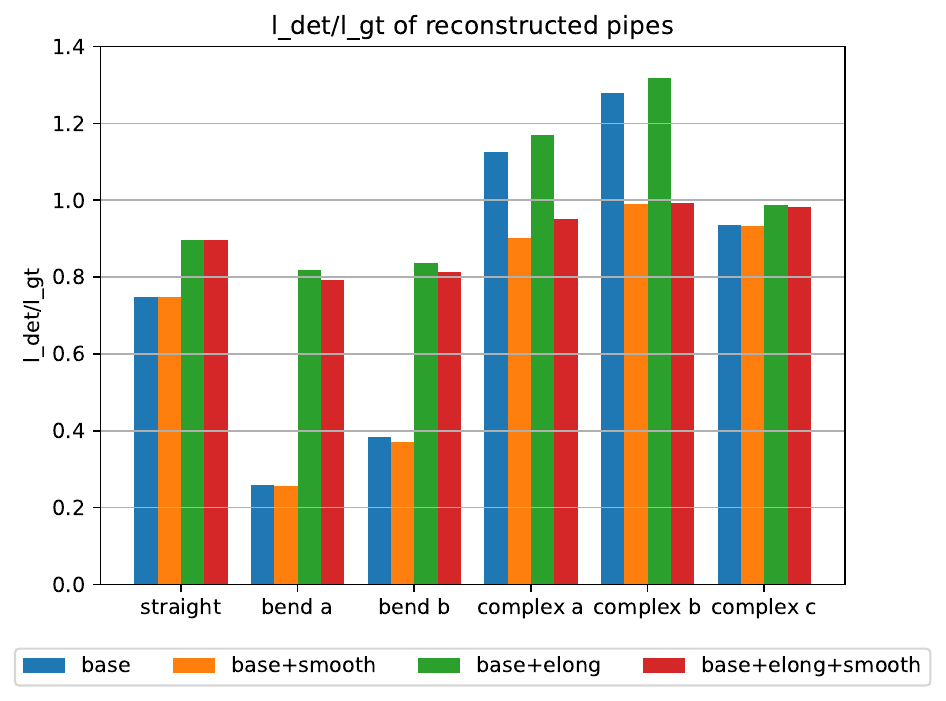}}
  \caption{example results for (a) IoU, (b) $\overline{r}$ ratio, (c) length ratio of reconstructed straight pipes, bends and more complex pipes}
  \label{fig:example_results} 
\end{figure*}

Exemplary reconstruction results are shown in Figure \ref{fig:example_results} (as charts) and Figure \ref{fig:example_results_pipe4_short} (as geometry). 
For further examples see Appendix \ref{appendix:A}.
For a straight pipe element smoothing has no significant effect, whereas elongation improves IoU and length ratio.
For the two bends, the IoU increases by more than 300~\%.
This is due to the elongation at both pipe ends, which can be also seen in the $l_{det}/l_{gt}$ ratio.
A longer pipe axis also improves the mean radius.

For the more complex pipe elements there are different effects caused by elongation and smoothing, depending if the base pipe length is under- or overestimated.
Where the base length is underestimated (complex c) smoothing and elongation have a comparable effect as for straight pipe elements.
For the pipes complex a and complex b, the base length is overestimated.
This results from a zigzag like skeleton, which particularly affects hull reconstruction, showing multiple spikes (Figures \ref{fig:example_results_pipe4_short}, \ref{fig:example_results_pipe3}, \ref{fig:example_results_pipe4}).
Due to smoothing the spikes are avoided, pipe length is no longer overestimated and the determination of the mean radius gets worse.

\begin{figure}[]
	\centering
	\subfloat[base + elong]{\includegraphics[width=0.5\linewidth]{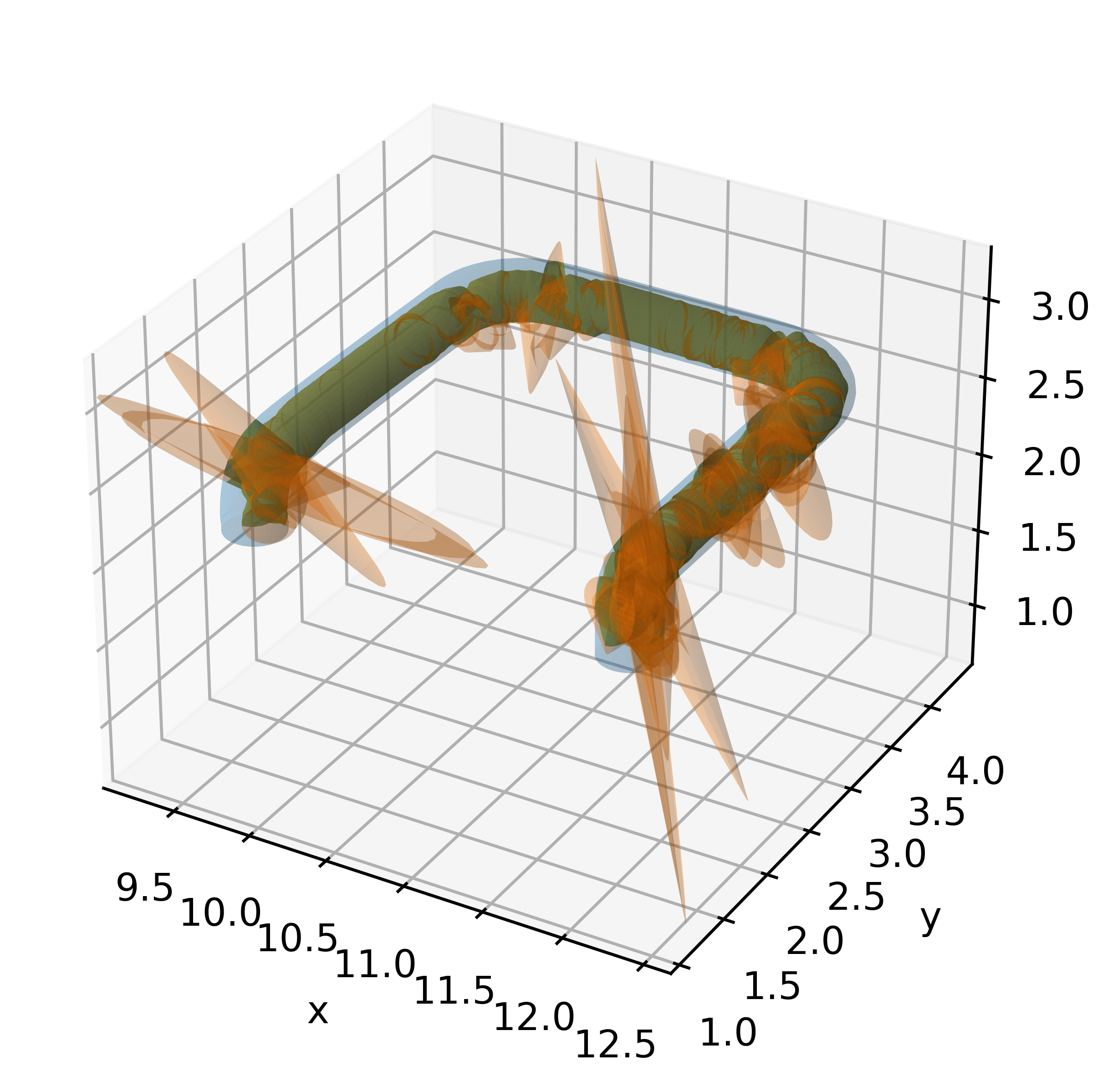}}
	\subfloat[base + elong + smooth]{\includegraphics[width=0.5\linewidth]{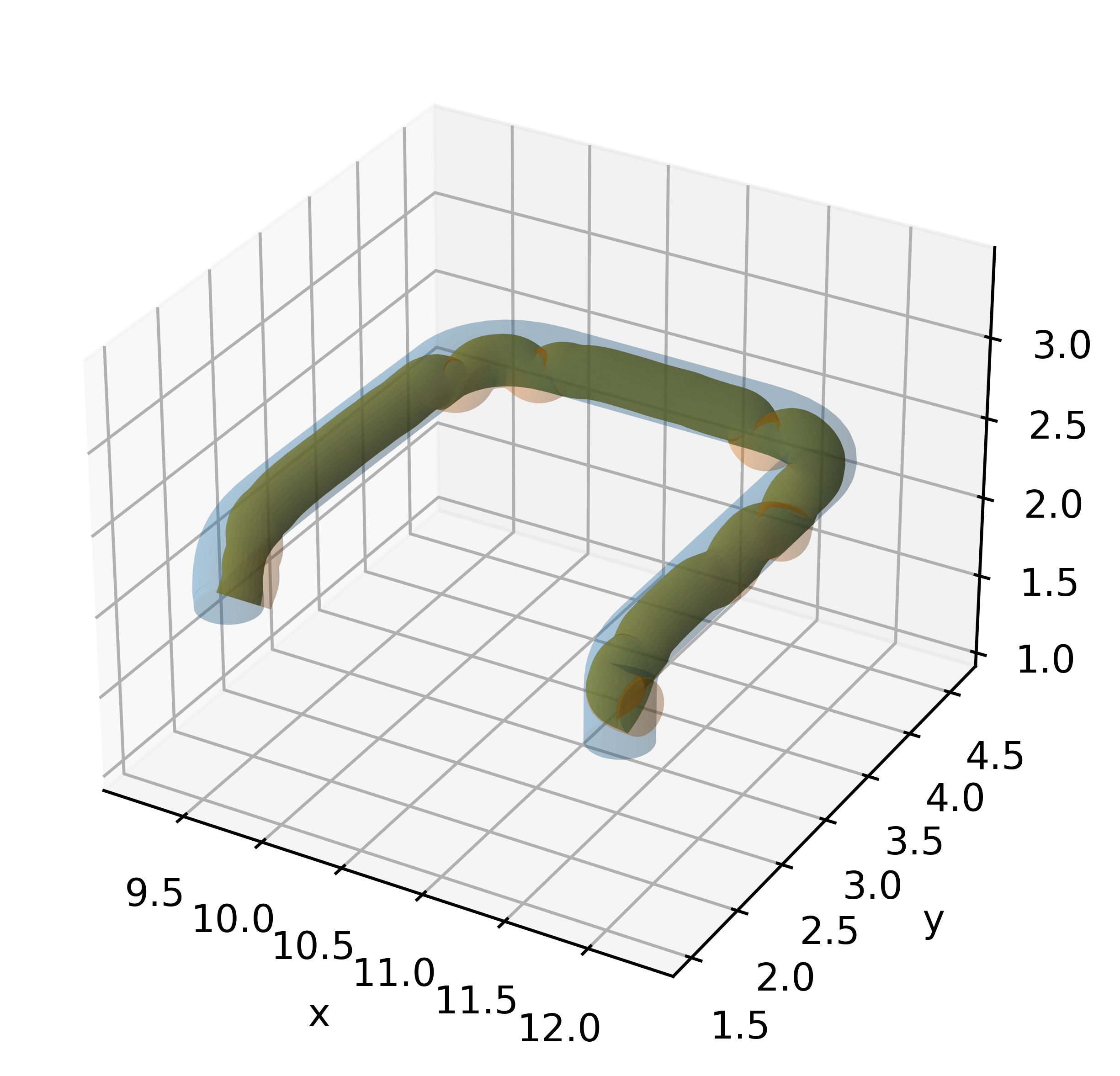}}
	\caption{pipe complex b; blue volume as ground truth, orange volume as reconstructed hull; $r_{gt}$=0.18~m, $l_{gt}$=7.85~m}
	\label{fig:example_results_pipe4_short}  
\end{figure}

IoU is influenced by errors in length and radius determination.
To provide insight into how these errors affect the IoU  several sample cases were explored. 
In case of two cylinders with the same central axis and same height but different radii, which is comparable for error in radius, the IoU is affected linearly proportional to $\Delta\text{IoU}\sim~2|\Delta r|/r$.
In case the radius is determined correct, but there is an error in height determination $\Delta\text{IoU}\sim~|\Delta h|/h$.
The rolling sphere algorithm shifts the skeleton axis closer to ground truth central axis. In case the central axis and the skeleton axis are eccentric IoU is affected by  
\begin{equation}
\label{IoU_eccentric}
\Delta \text{IoU} = 1-\frac{\text{cos}^{-1}(\frac{d}{2r})-\frac{d}{2r}\sqrt{1-(\frac{d}{2r})^2}}{\pi-\text{cos}^{-1}(\frac{d}{2r})-\frac{d}{2r}\sqrt{1-(\frac{d}{2r})^2}}.
\end{equation}
The eccentricity is described by $d=\sqrt{\Delta x^2+\Delta y^2}$. Assuming an error of $d = 0.01r$ results in $\Delta \text{IoU} \approx 0.01$.

\begin{table}
	\caption{Results after applying RDP algorithm} \label{tab:RDP}
	\centering
	\begin{tabular}{lrrr}
		&  IoU & $l/l_{gt}$ & $N_{RDP}/N$ \\
		\hline\\[-0.7em]
		base skeleton & 0.5527 & 0.7025 & 0.0670  \\
		+ smooth & 0.5468 & 0.6985 & 0.0122  \\
		+ elong & 0.6407 & 0.8450 & 0.0544  \\
		+ elong + smooth & 0.6322 & 0.8430 & 0.0131 \\
		\hline
	\end{tabular}
\end{table}

If RDP is added to the reconstruction process, the values for IoU and length ration remain quite similar, but the amount $N$ of describing spline points  is significantly reduced to 1-5~\% (Table~\ref{tab:RDP}).
These are for some example pipes still more points than needed to describe the ground truth spline in UE, but it is in the same order of magnitude. 

\section{Conclusion}
This paper demonstrated the feasibility of reconstructing pipes from incomplete point cloud data, showcasing promising results in the field of automated model generation.
To further validate and refine this process, it is essential to conduct testing of the reconstruction algorithm on real-world point clouds.
While traditional building settings usually have pipes concealed behind walls and industrial environments often have the space to lay out pipes in clear and open patterns, the maritime domain pipe networks are frequently exposed but tightly packed to maximise the space usage in restricted spaces, making them accessible but still difficult to analyse and understand.
Compact and complex environments such as e.~g.\ ship engine rooms complicate the task of detecting and reconstructing pipe networks.
Doing this manually is a challenging, time-consuming and also error prone task, making automation a necessity.

One crucial element that remains to be integrated in the algorithm is a proper handling of junctions, as they are not yet included in the dataset.
Except for this last part, the reconstructed pipes could be stored in a compact data representation, and can also be effectively modelled in UE using a component catalogue. 
This enables the creation of realistic and accurate digital twins of pipe systems which could be used for a variety of applications, including simulation, analysis, and predictive maintenance, contributing to the reliability and sustainability of maritime infrastructures and vessels.

\vspace{1em}

\ifdefined\MARESEC
	For more result images see \url{https://arxiv.org/todo}.
\fi

\ifdefined\ARXIV

	\clearpage
	\appendix
	\section*{Exemplary Results of Reconstructed Pipes}\label{appendix:A}
	\begin{figure*}[!hb]
		\vspace{-15em}
		\centering
		\subfloat[base]{\includegraphics[width=0.23\linewidth]{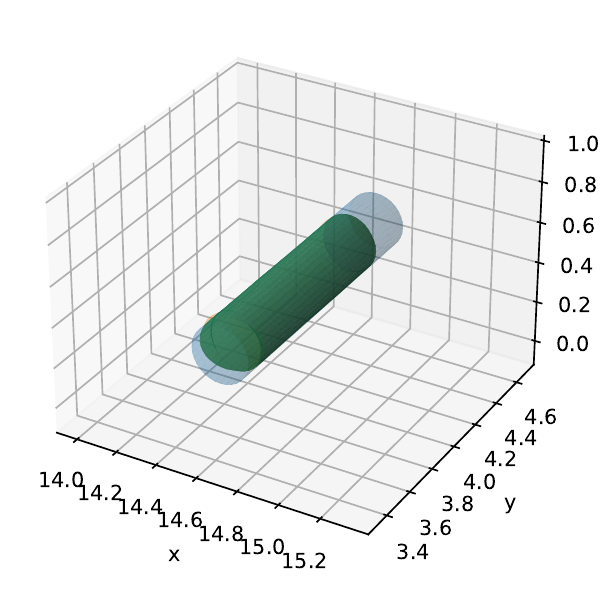}}
		\subfloat[base + smooth]{\includegraphics[width=0.23\linewidth]{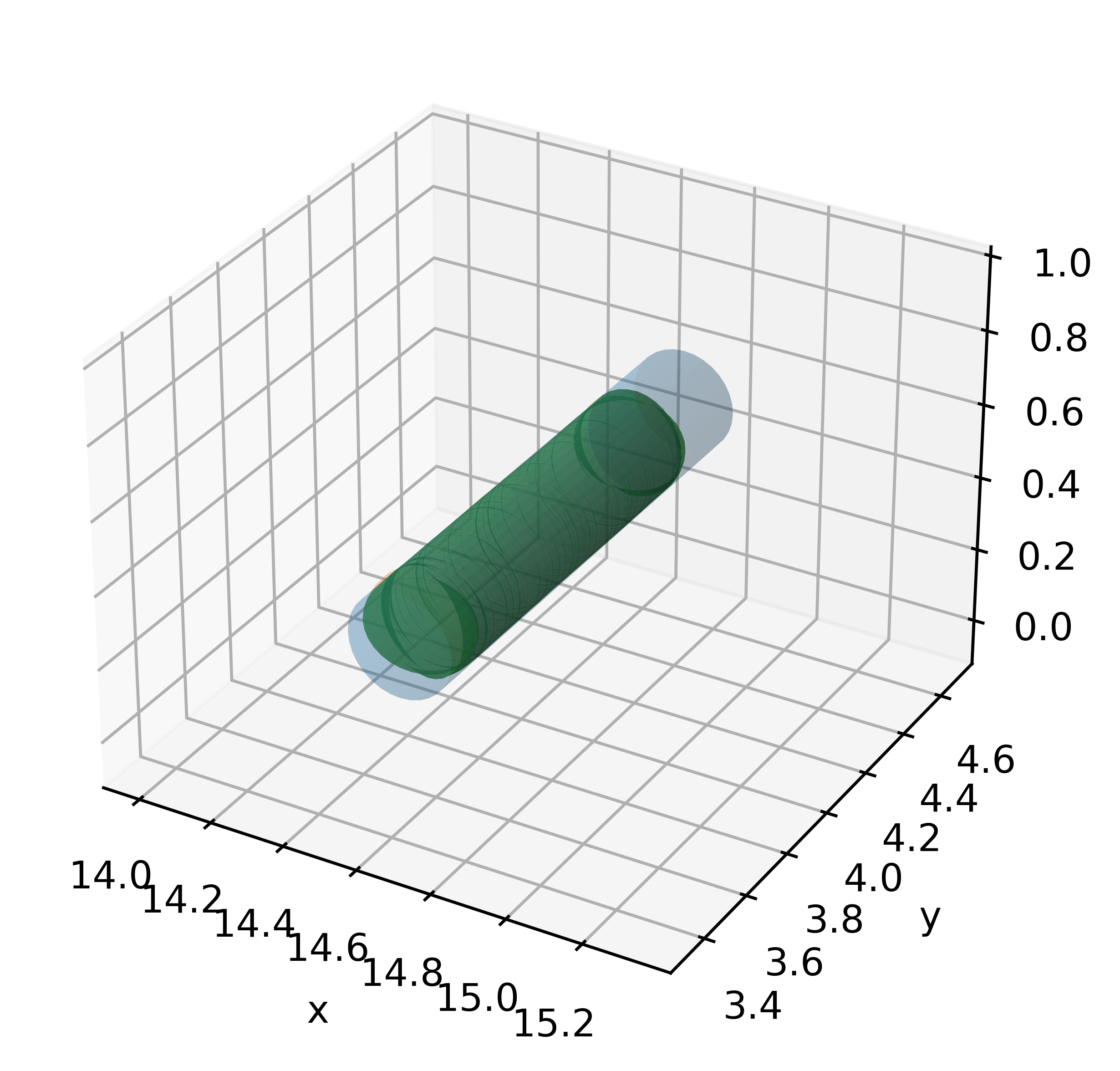}}
		\subfloat[base + elong]{\includegraphics[width=0.23\linewidth]{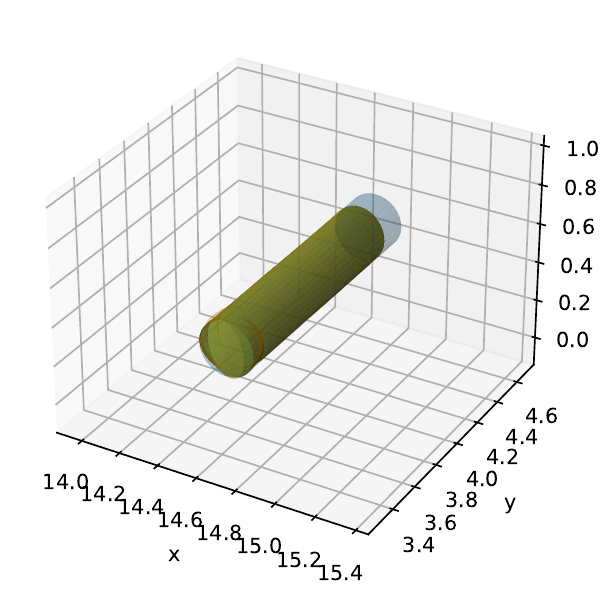}}
		\subfloat[base + elong + smooth]{\includegraphics[width=0.23\linewidth]{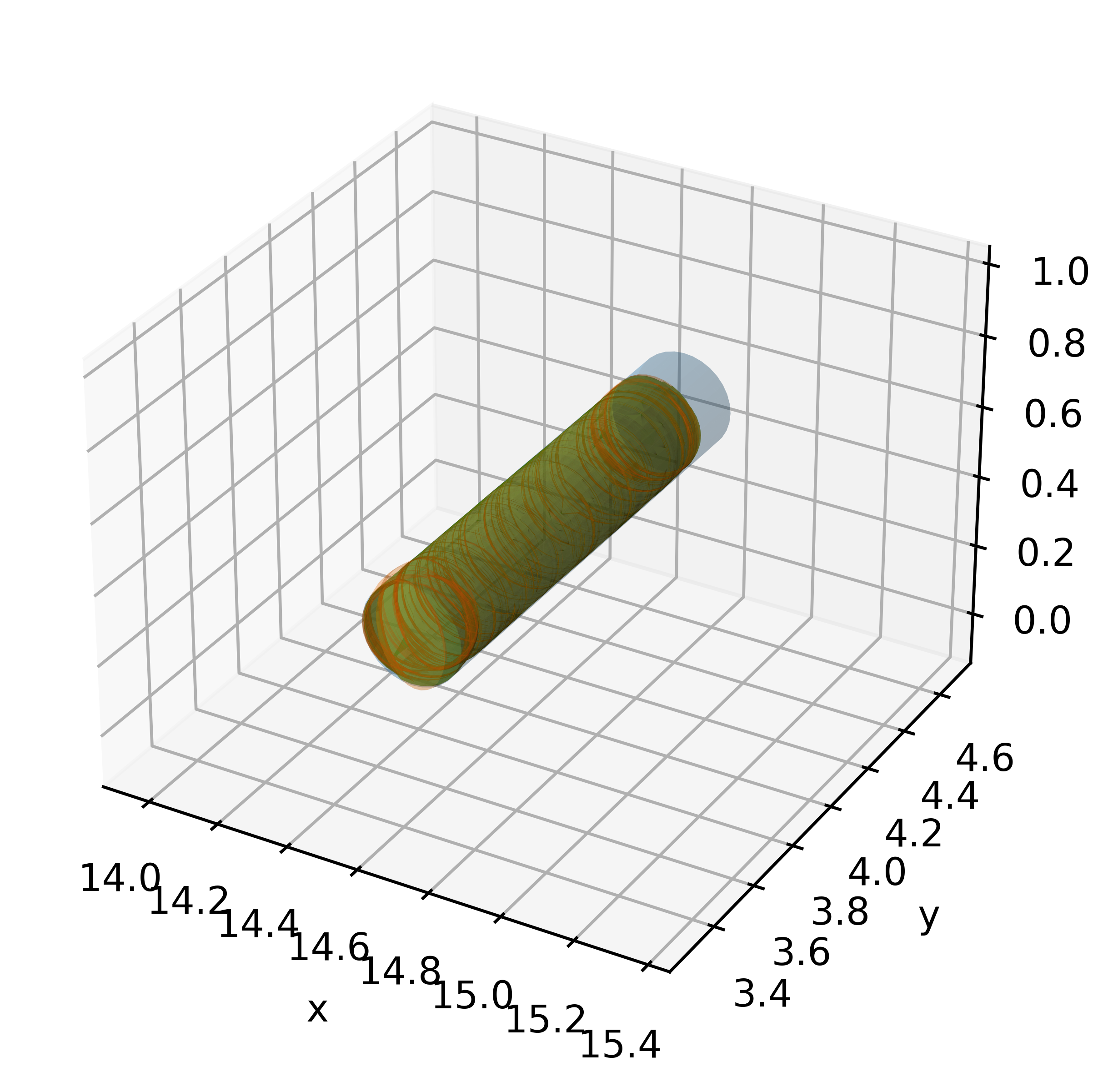}}
		\caption{ example results for a straight pipe; $r_{gt}$=0.13~m, $l_{gt}$=1.33~m; blue volume is ground truth, orange volume is reconstructed, green volume is intersection}
		\label{fig:example_results_pipe7} 
	\end{figure*}
	
	\begin{figure*}[!b]
		\centering
		\subfloat[base]{\includegraphics[width=0.23\linewidth]{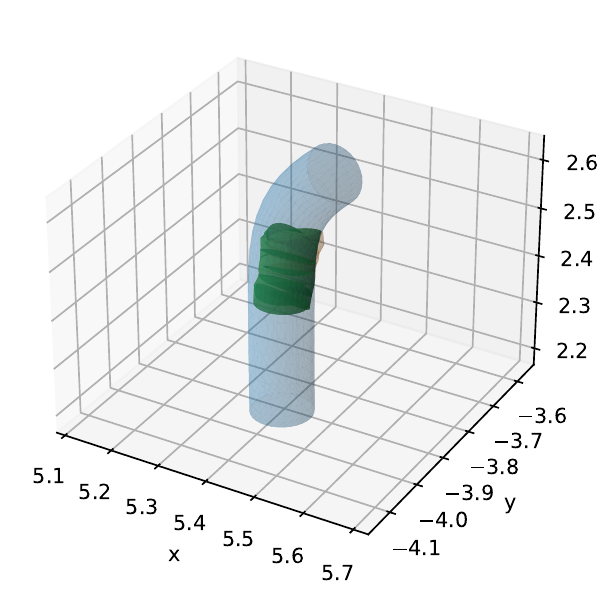}}
		\subfloat[base + smooth]{\includegraphics[width=0.23\linewidth]{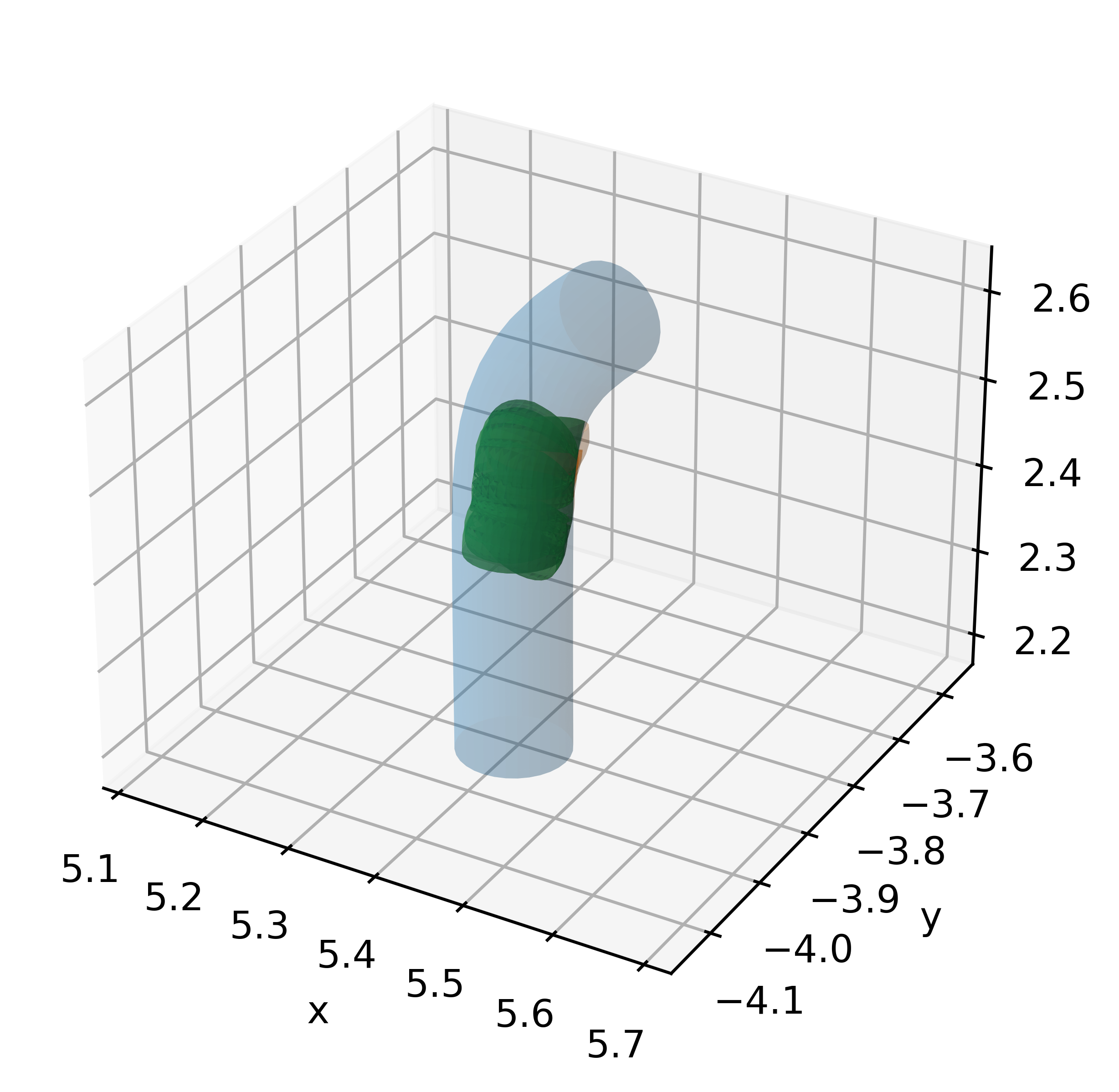}}
		\subfloat[base + elong]{\includegraphics[width=0.23\linewidth]{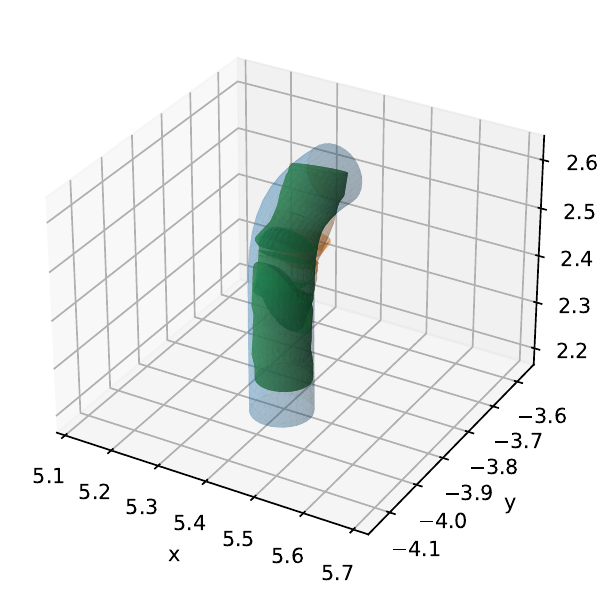}}
		\subfloat[base + elong + smooth]{\includegraphics[width=0.23\linewidth]{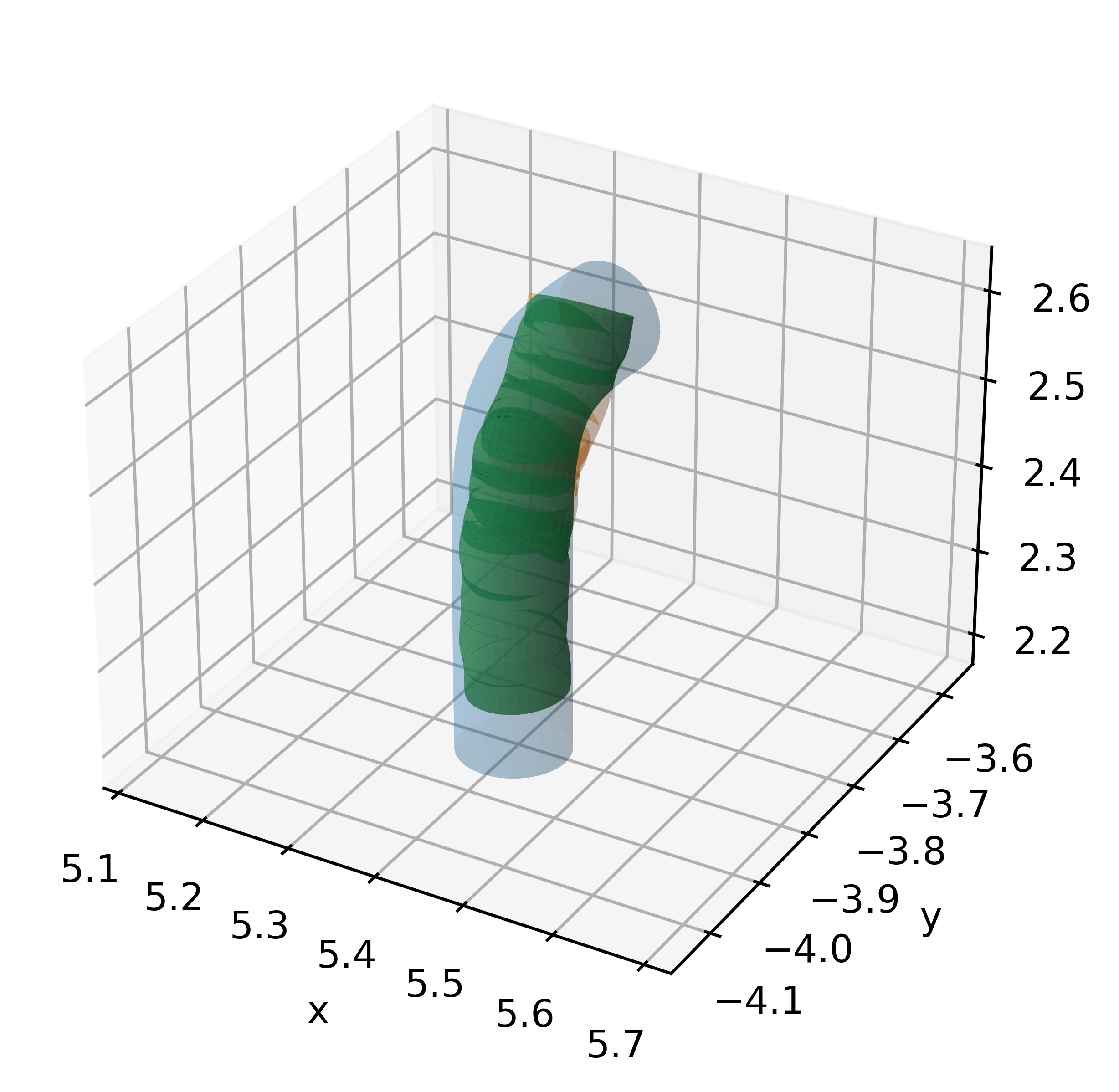}}
		\caption{example results for bend a; $r_{gt}$=0.06~m, $l_{gt}$=0.54~m; blue volume is ground truth, orange volume is reconstructed, green volume is intersection}
		\label{fig:example_results_pipe14} 
	\end{figure*}
	
	\begin{figure*}[!b]
		\centering
		\subfloat[base]{\includegraphics[width=0.23\linewidth]{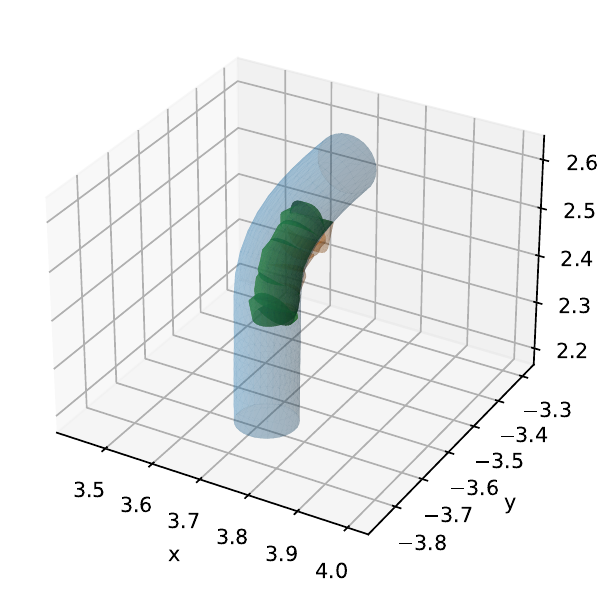}}
		\subfloat[base + smooth]{\includegraphics[width=0.23\linewidth]{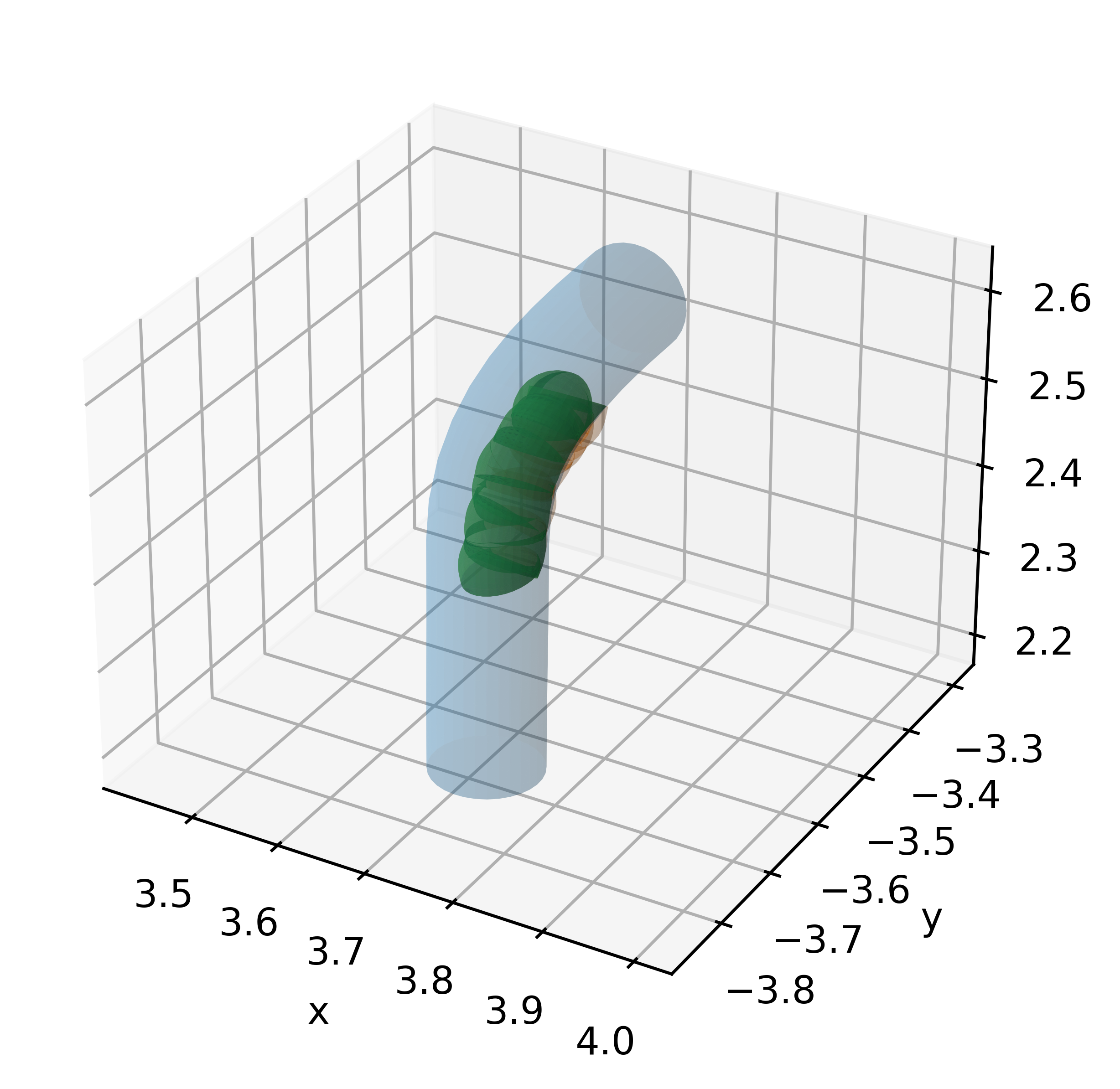}}
		\subfloat[base + elong]{\includegraphics[width=0.23\linewidth]{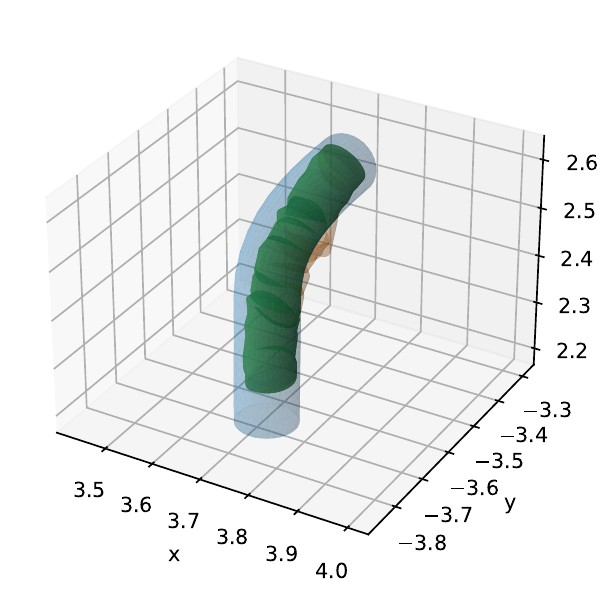}}
		\subfloat[base + elong + smooth]{\includegraphics[width=0.23\linewidth]{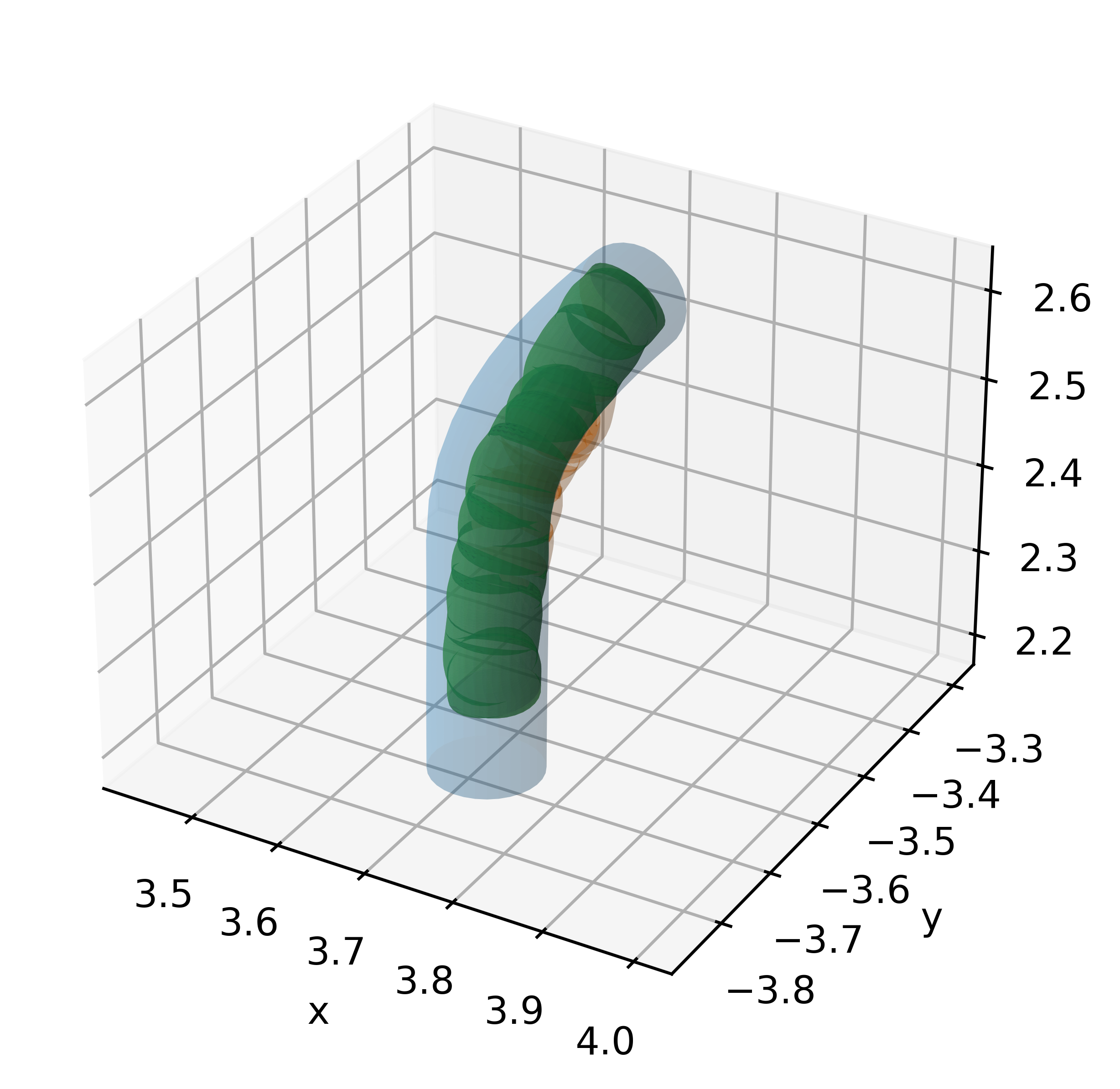}}
		\caption{example results for bend b; $r_{gt}$=0.06~m, $l_{gt}$=0.61~m; blue volume is ground truth, orange volume is reconstructed, green volume is intersection}
		\label{fig:example_results_pipe16} 
	\end{figure*}
	
	\begin{figure*}[!b]
		\centering
		\subfloat[base]{\includegraphics[width=0.23\linewidth]{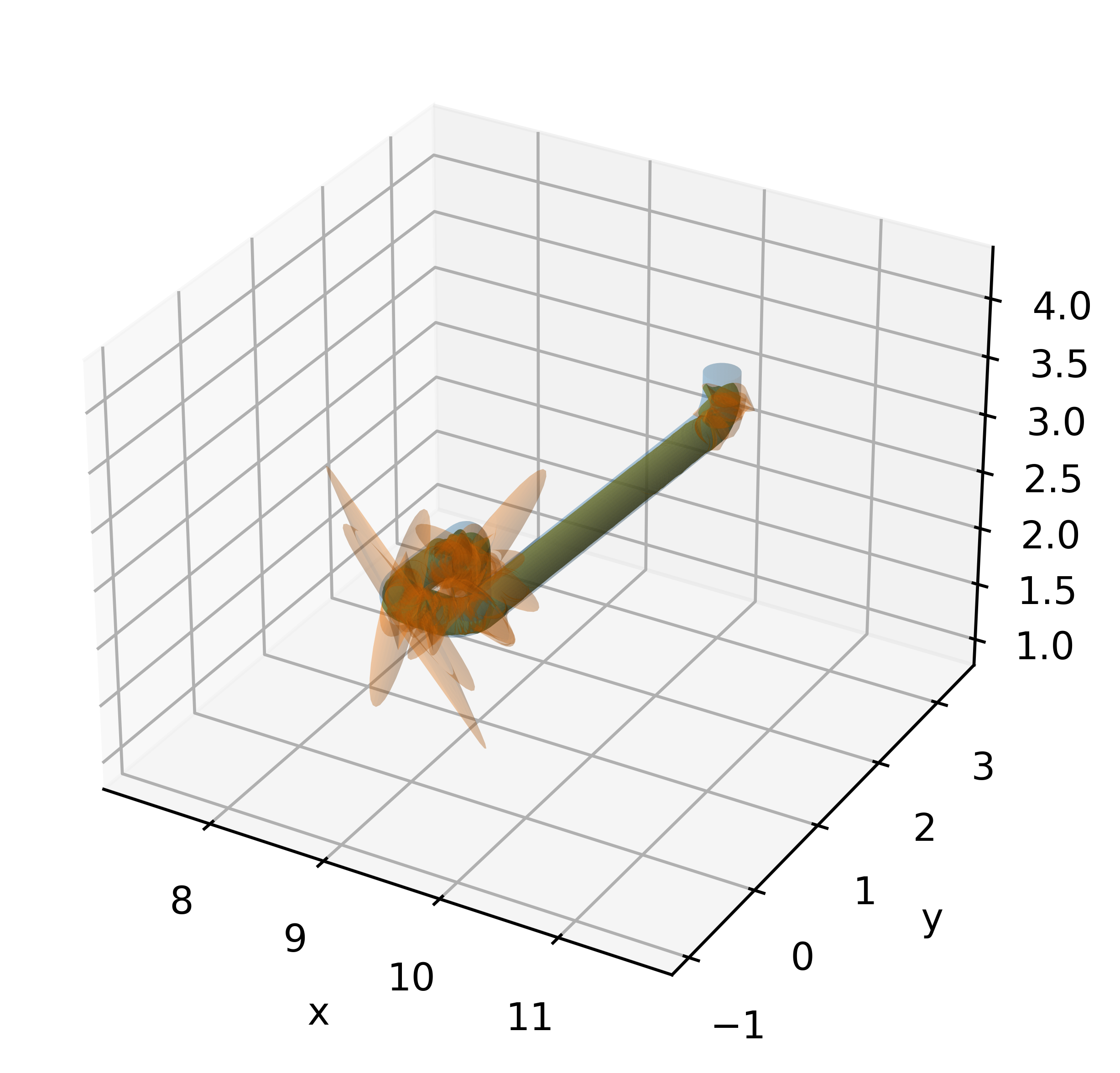}}
		\subfloat[base + smooth]{\includegraphics[width=0.23\linewidth]{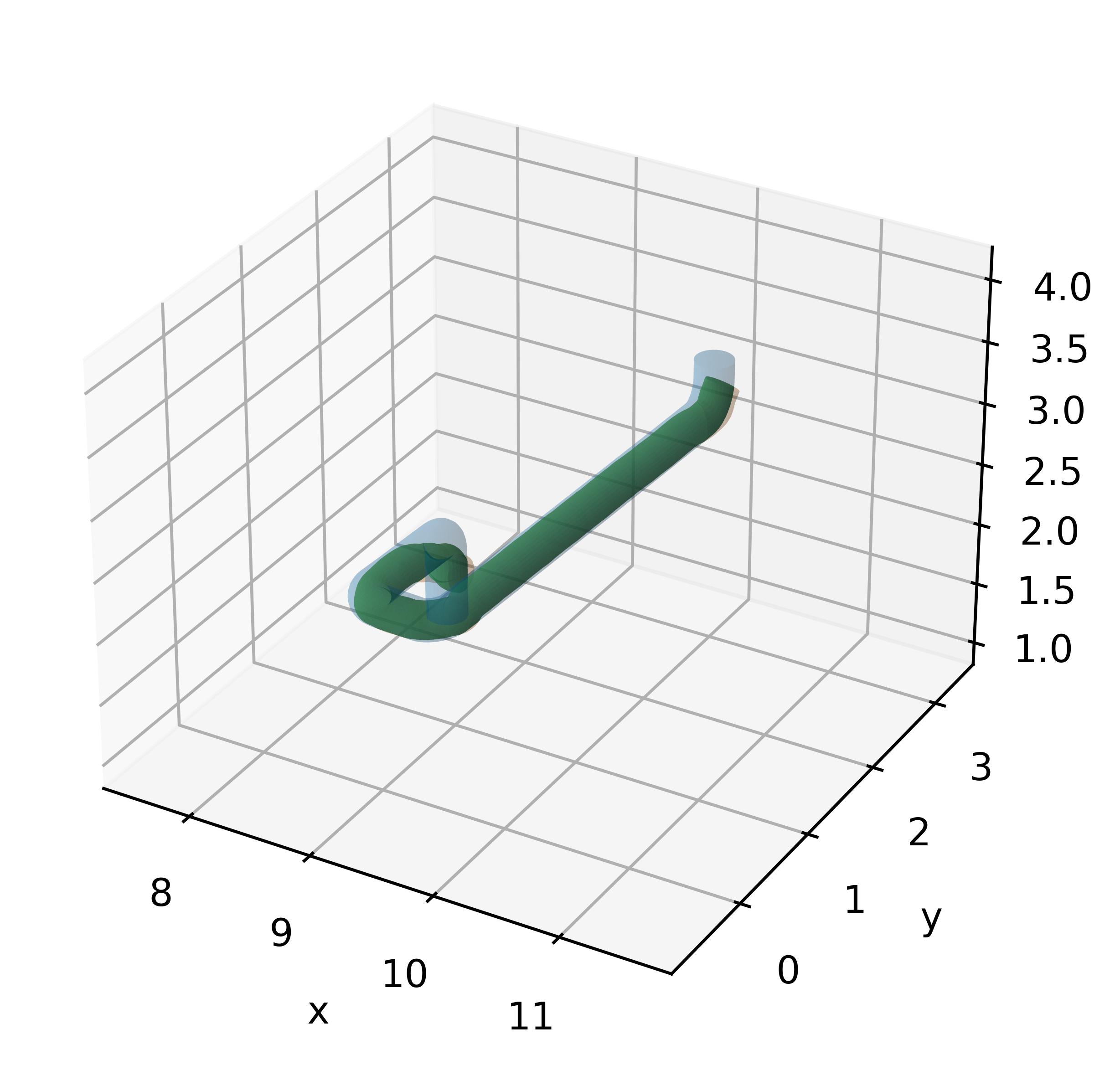}}
		\subfloat[base + elong]{\includegraphics[width=0.23\linewidth]{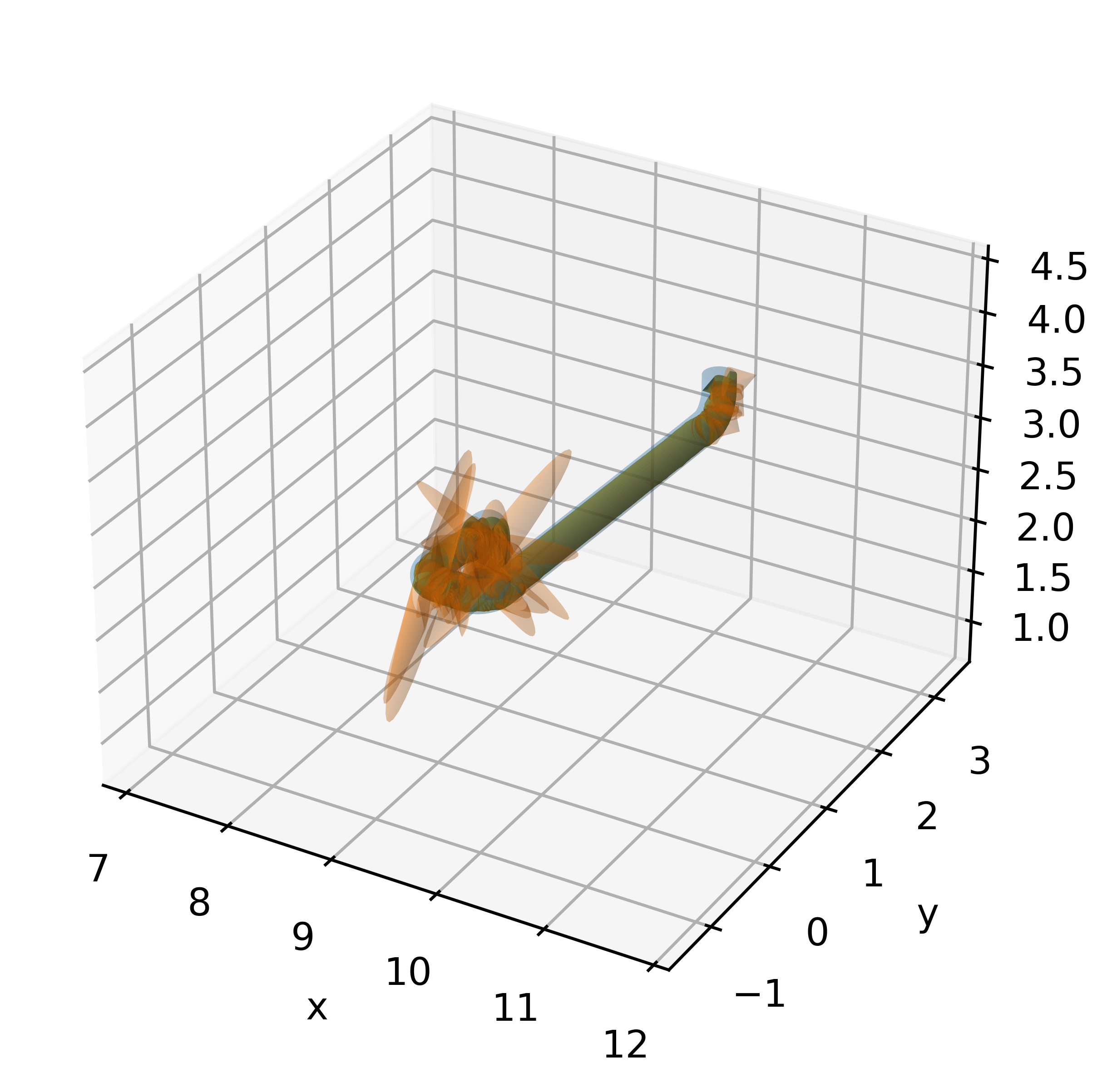}}
		\subfloat[base + elong + smooth]{\includegraphics[width=0.23\linewidth]{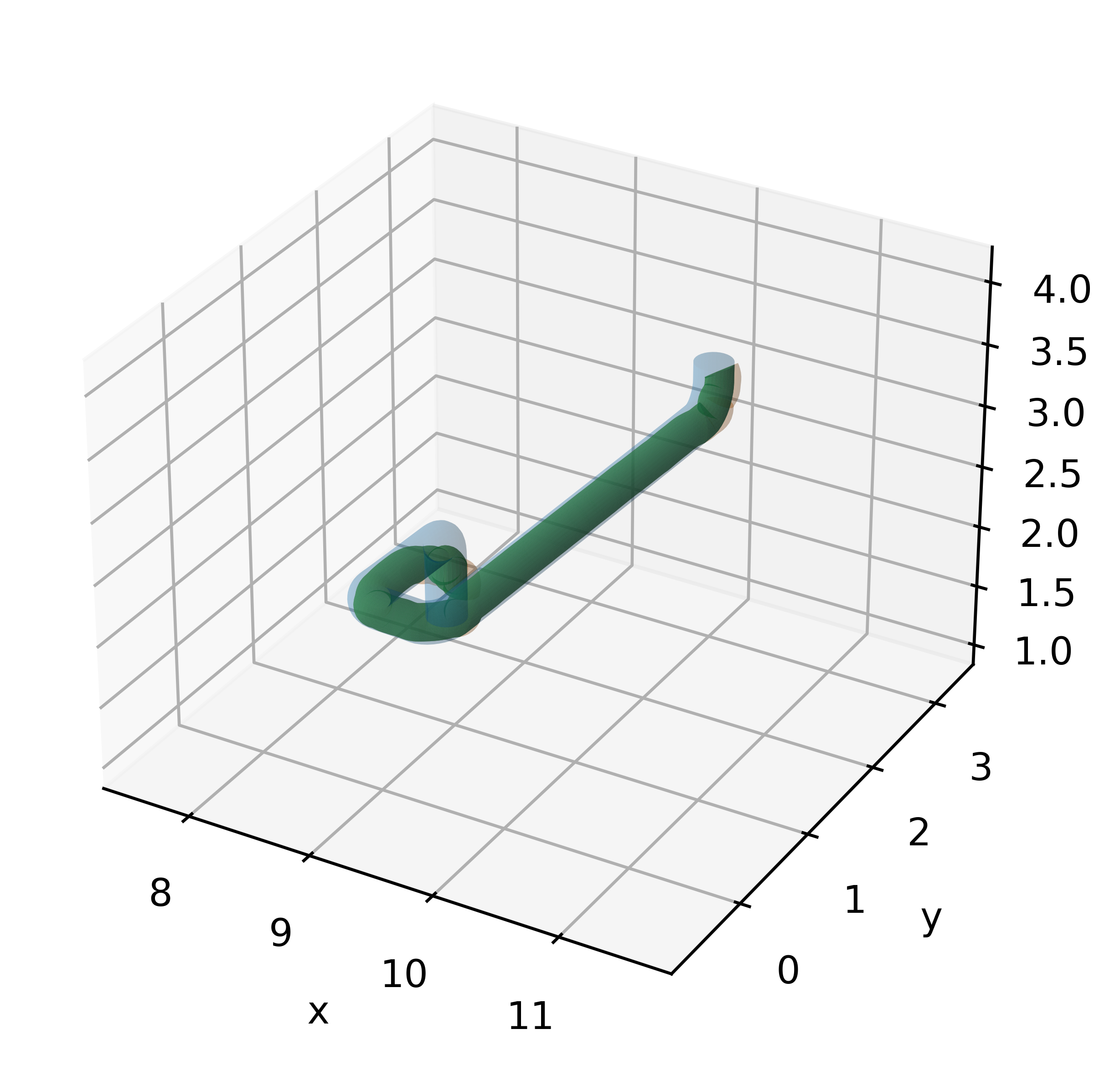}}
		\caption{example results for pipe complex a; $r_{gt}$=0.15~m, $l_{gt}$=6.15~m;  blue volume is ground truth, orange volume is reconstructed, green volume is intersection}
		\label{fig:example_results_pipe3} 
	\end{figure*}
	
	\begin{figure*}[!b]
		\centering
		\subfloat[base]{\includegraphics[width=0.23\linewidth]{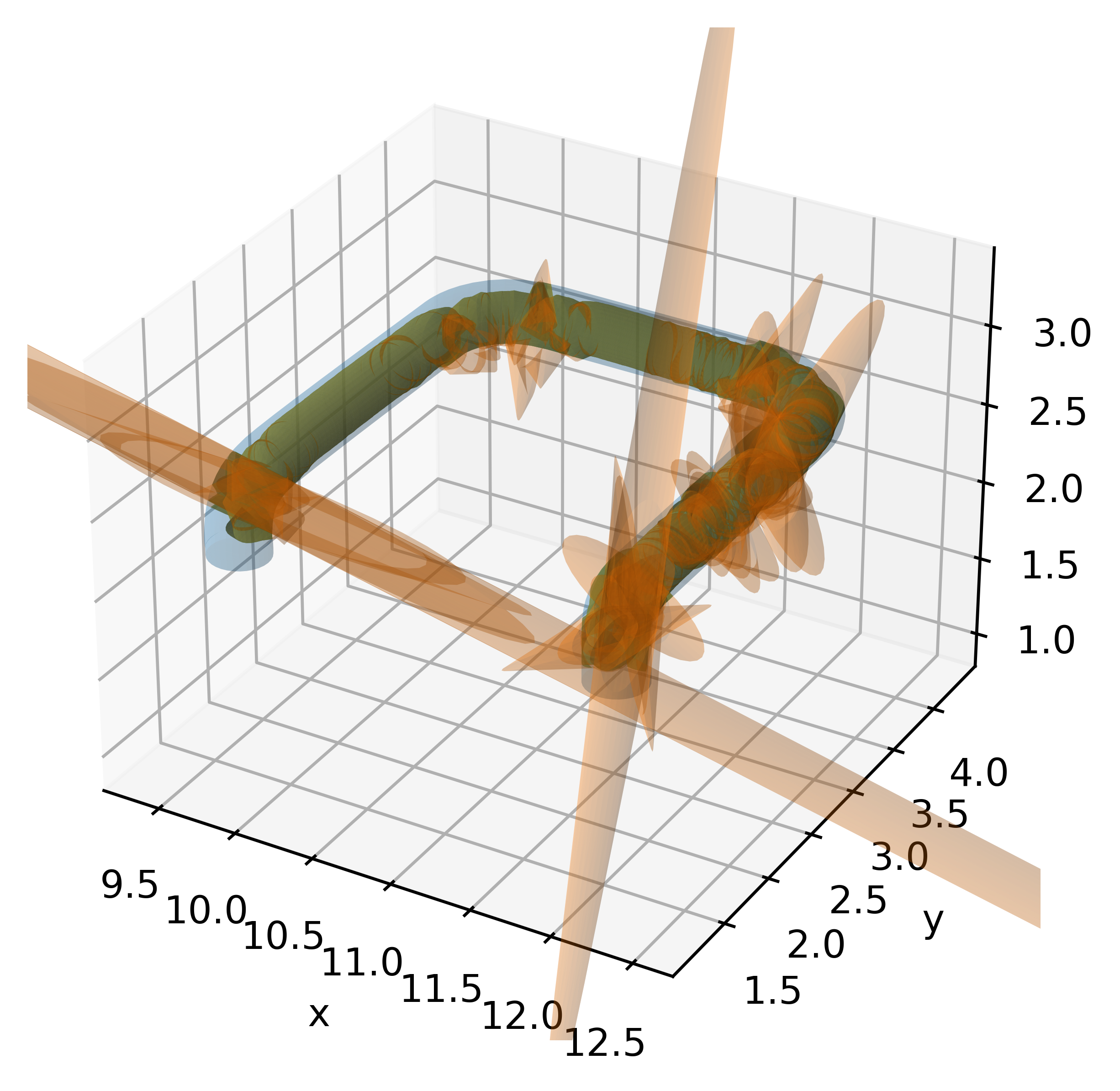}}
		\subfloat[base + smooth]{\includegraphics[width=0.23\linewidth]{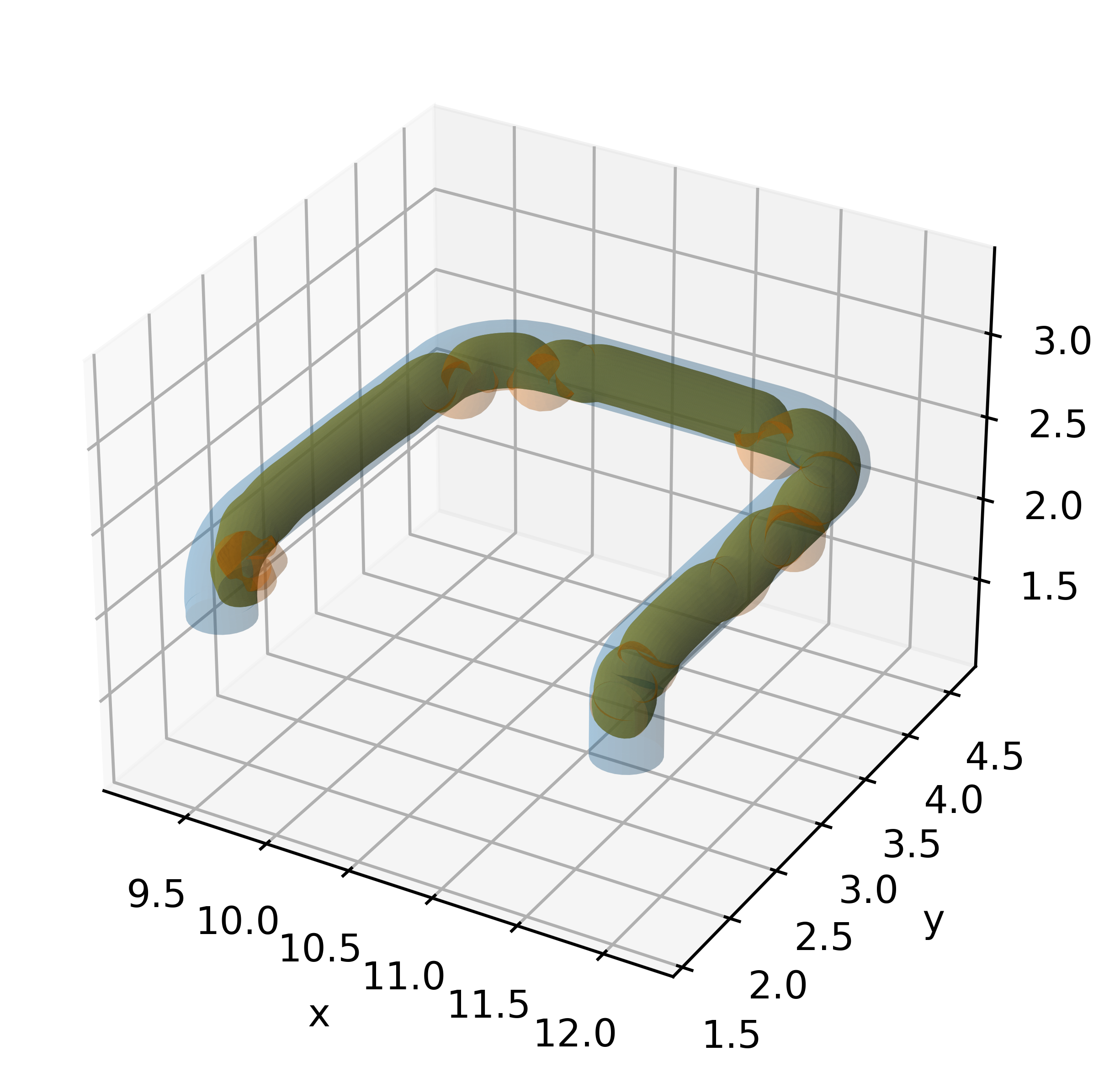}}
		\subfloat[base + elong]{\includegraphics[width=0.23\linewidth]{Figures/Intersection_elong_Pipe14.png}}
		\subfloat[base + elong + smooth]{\includegraphics[width=0.23\linewidth]{Figures/Intersection_elong_smooth_Pipe14.png}}
		\caption{ example results for pipe complex b; $r_{gt}$=0.18~m, $l_{gt}$=7.85~m; blue volume is ground truth, orange volume is reconstructed, green volume is intersection}
		\label{fig:example_results_pipe4} 
	\end{figure*}
	
	\begin{figure*}[!b]
		\centering
		\subfloat[base]{\includegraphics[width=0.23\linewidth]{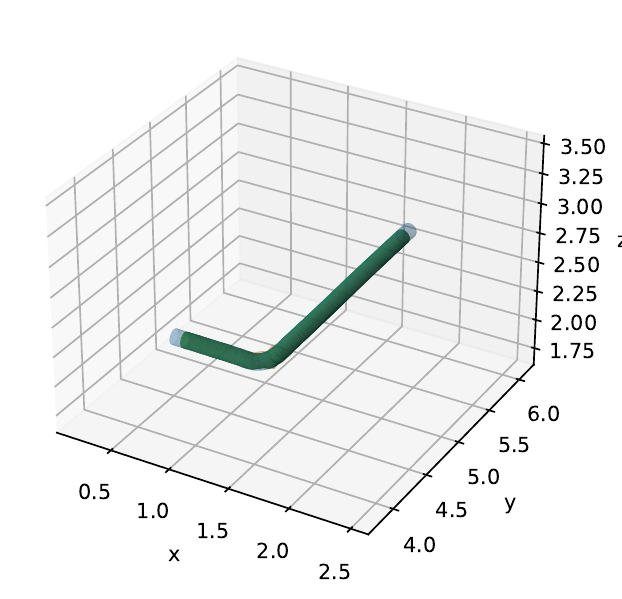}}
		\subfloat[base + smooth]{\includegraphics[width=0.23\linewidth]{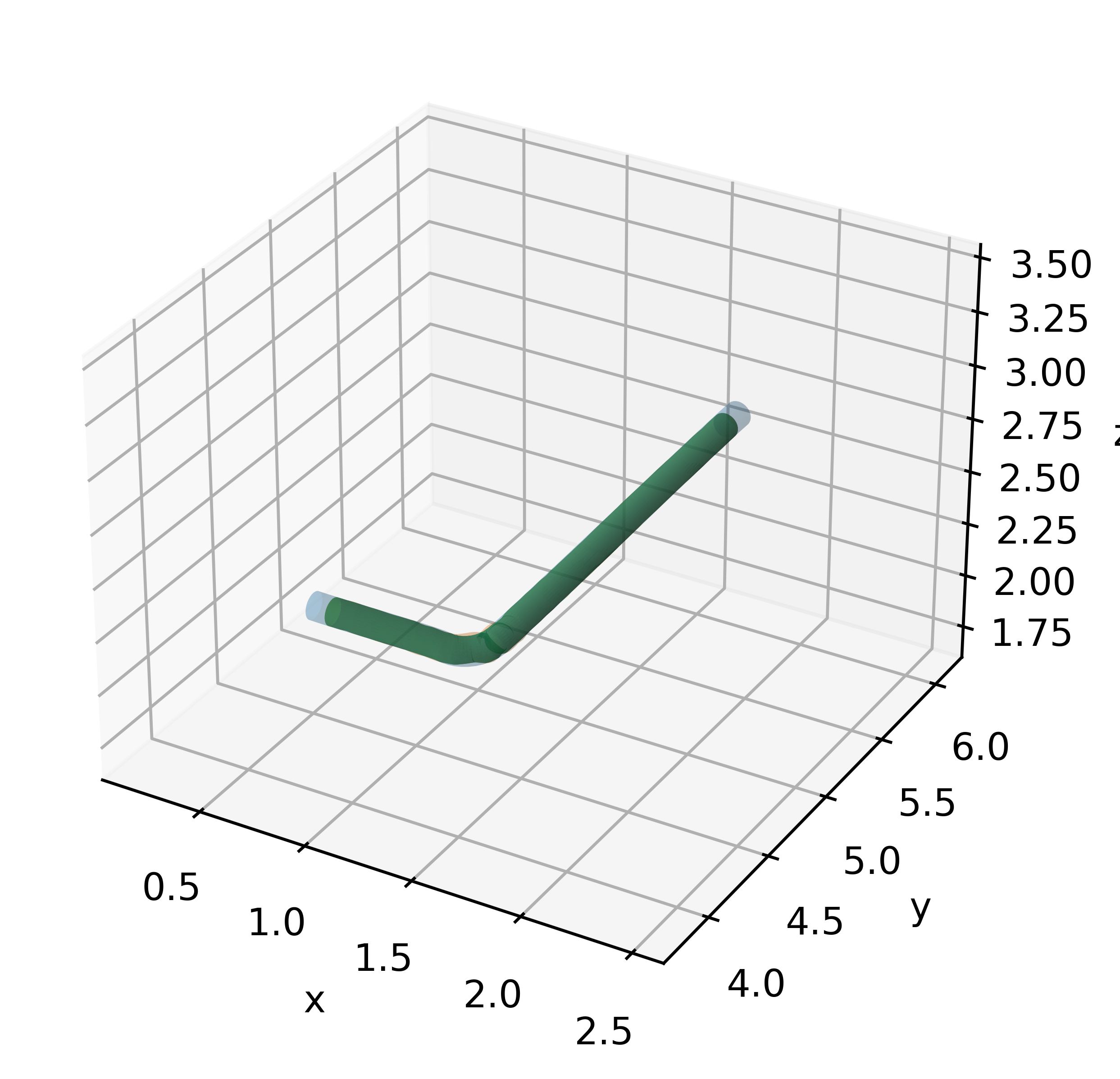}}
		\subfloat[base + elong]{\includegraphics[width=0.23\linewidth]{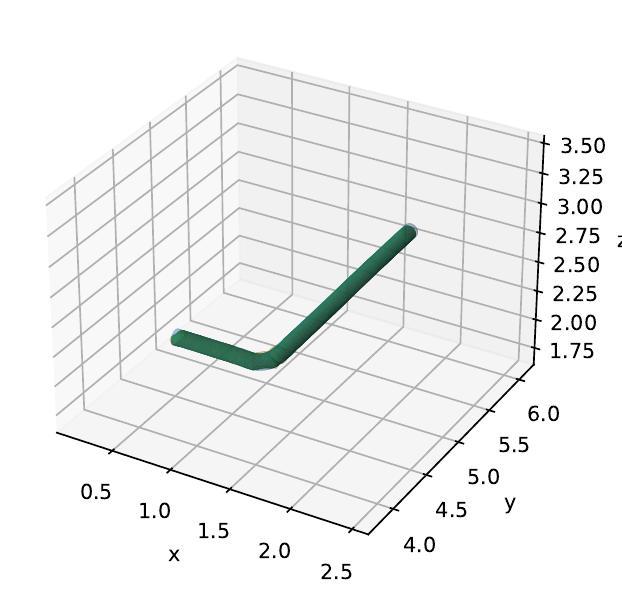}}
		\subfloat[base + elong + smooth]{\includegraphics[width=0.23\linewidth]{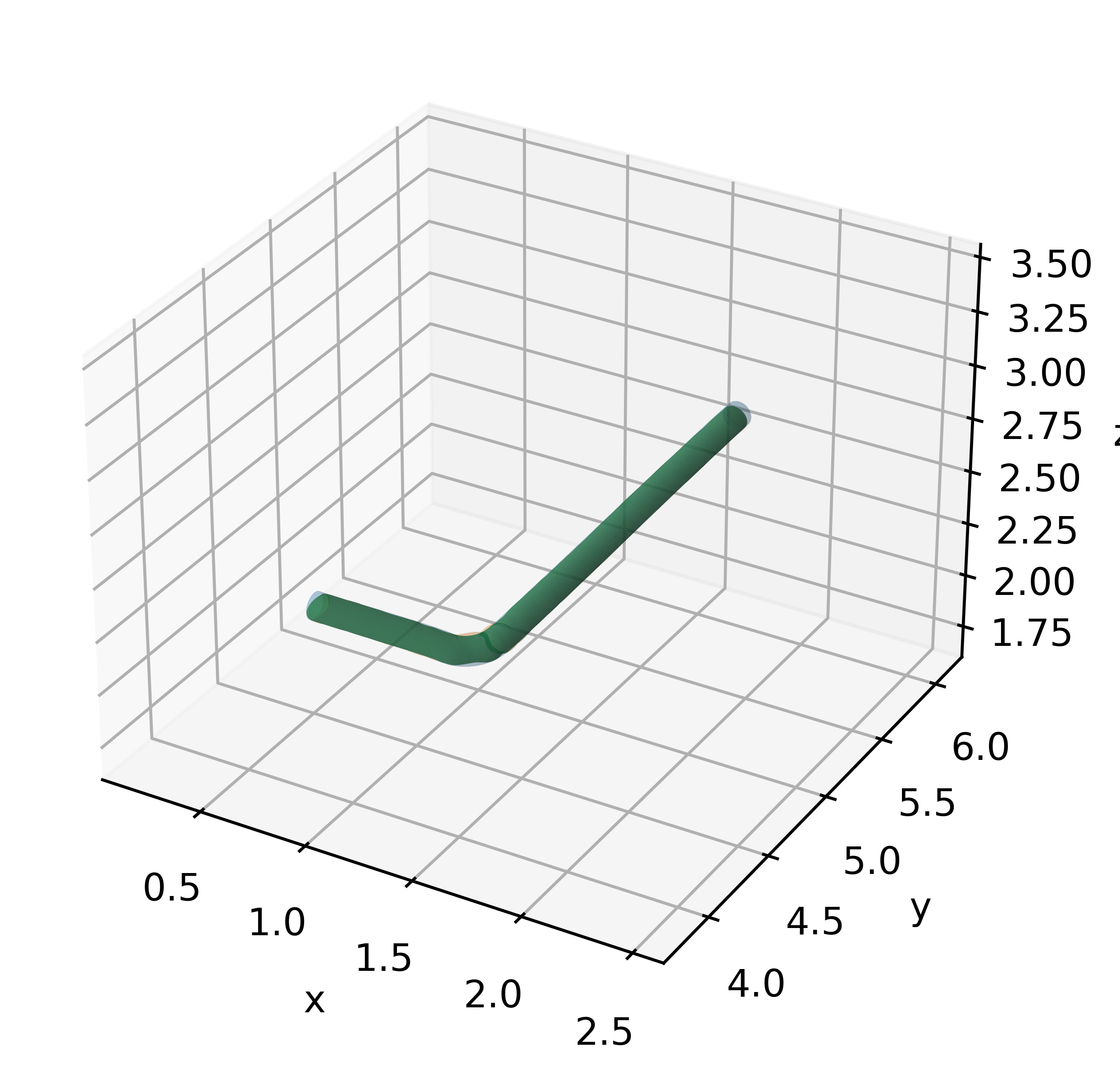}}
		\caption{example results for pipe complex c; $r_{gt}$=0.06~m, $l_{gt}$=2.92~m; blue volume is ground truth, orange volume is reconstructed, green volume is intersection}
		\label{fig:example_results_pipe38} 
	\end{figure*}

\fi

\end{document}